\documentclass[sigconf]{aamas}
\usepackage{balance}

\usepackage{times}
\usepackage{soul}
\usepackage{url}
\usepackage[utf8]{inputenc}
\usepackage{graphicx}
\usepackage{amsmath}
\usepackage{amsthm}
\usepackage{booktabs}
\usepackage{algorithmicx}
\usepackage{algpseudocode}
\usepackage{hyphenat}
\usepackage{siunitx}
\algnewcommand{\IIf}[1]{\State\algorithmicif\ #1\ \algorithmicthen}
\algnewcommand{\IEndIf}{\unskip\ \algorithmicend\ \algorithmicif}

\usepackage[textsize=scriptsize,colorinlistoftodos]{todonotes}

\newtheorem{definition}{Definition}

\newcommand*{\textcite}{\citet}

\usepackage{xspace}

\newcommand*{\golog}{\textsc{Golog}\xspace}
\newcommand*{\congolog}{\textsc{ConGolog}\xspace}

\newcommand*{\es}{\ensuremath{\mathcal{E \negthinspace S}}\xspace}
\newcommand*{\esg}{\ensuremath{\mathcal{E \negthinspace S \negthinspace G}}\xspace}
\newcommand*{\ds}{\texorpdfstring{\ensuremath{\mathcal{D \negthinspace S}}}{DS}\xspace}
\newcommand*{\dsg}{\texorpdfstring{\ensuremath{\mathcal{D \negthinspace S \negthinspace G}}}{DSG}\xspace}
\newcommand*{\bhl}{BHL\xspace}

\newcommand*{\poss}{\ensuremath{\mathit{Poss}}}
\newcommand*{\exec}{\ensuremath\mathit{exec}}
\newcommand*{\eqdef}{\ensuremath{:=}}

\newcommand*{\equivspace}{\ensuremath\,\equiv\;}

\newcommand*{\belconnector}{\ensuremath{\mathbf{B}}\xspace}
\newcommand*{\obelconnector}{\ensuremath{\mathbf{O}}\xspace}
\newcommand*{\bel}[2]{\ensuremath{\belconnector\mleft(#1\,\mathbf{:}\,#2\mright)}}

\newcommand*{\know}[1]{\ensuremath{\mathbf{K}#1}}

\newcommand*{\tnext}[2][]{\ensuremath{\mathbf{X}^{#1}_{#2}}\if#1{\,}\fi}
\newcommand*{\tprev}[1]{\ensuremath{\mathbf{V}_{#1}}\if#1{\,}\fi}
\newcommand*{\fut}[1]{\ensuremath{\mathbf{F}_{#1}}\if#1{\,}\fi}
\newcommand*{\past}[1]{\ensuremath{\mathbf{P}_{#1}}\if#1{\,}\fi}
\newcommand*{\glob}[1]{\ensuremath{\mathbf{G}_{#1}}\if#1{\,}\fi}
\newcommand*{\hist}[1]{\ensuremath{\mathbf{H}_{#1}}\if#1{\,}\fi}
\newcommand*{\mi}[1]{\ensuremath{\mathit{#1}}}
\newcommand*{\la}{\left\langle}
\newcommand*{\ra}{\right\rangle}
\mathchardef\mhyphen="2D 

\newcommand*{\near}{\ensuremath{\mathit{near}}\xspace}
\newcommand*{\midpos}{\ensuremath{\mathit{mid}}\xspace}
\newcommand*{\far}{\ensuremath{\mathit{far}}\xspace}
\newcommand*{\goto}{\ensuremath{\mathit{goto}}\xspace}
\newcommand*{\sigmah}{\ensuremath{\Sigma_{\mi{goto}}}\xspace}
\newcommand*{\sigmal}{\ensuremath{\Sigma_{\mi{move}}}\xspace}
\newcommand*{\loc}{\ensuremath{\mathit{Loc}}\xspace}
\newcommand*{\at}{\ensuremath{\mathit{At}}\xspace}
\newcommand*{\move}{\ensuremath{\mathit{move}}\xspace}

\newcommand*{\sonar}{\ensuremath{\mathit{sonar}}\xspace}

\newcommand*{\gspace}{\,}
\newcommand*{\gwhile}{\ensuremath{\gspace\mathbf{while}\gspace}}
\newcommand*{\gdo}{\ensuremath{\gspace\mathbf{do}\gspace}}
\newcommand*{\gdone}{\ensuremath{\gspace\mathbf{done}\gspace}}
\newcommand*{\gif}{\ensuremath{\gspace\mathbf{if}\gspace}}
\newcommand*{\gelif}{\ensuremath{\gspace\mathbf{elif}\gspace}}
\newcommand*{\gelse}{\ensuremath{\gspace\mathbf{else}\gspace}}
\newcommand*{\gthen}{\ensuremath{\gspace\mathbf{then}\gspace}}
\newcommand*{\gfi}{\ensuremath{\gspace\mathbf{fi}\gspace}}

\newcommand*{\traces}{\ensuremath{\mathcal{Z}}\xspace}
\newcommand*{\rigid}{\ensuremath{\mathcal{R}}\xspace}
\newcommand*{\atoms}{\ensuremath{\mathcal{P}}\xspace}
\newcommand*{\worlds}{\ensuremath{\mathcal{W}}\xspace}

\newcommand*{\states}{\ensuremath{\mathcal{S}}\xspace}
\newcommand*{\stateset}{\ensuremath{S}\xspace}

\newcommand*{\oicomp}{\ensuremath{\approx_{\textrm{oi}}}\xspace}
\newcommand*{\oi}{\ensuremath{\mathit{oi}}\xspace}
\newcommand*{\oisim}[1][w]{\sim_{#1}}
\newcommand*{\whfull}[2]{\ensuremath{\stateset^{#1}_{#2}}}
\newcommand*{\wh}[2][]{\whfull{e_h, w_h#1, z_h#1}{#2}}
\newcommand*{\wlfull}[2]{\ensuremath{\stateset^{#1}_{\ifthenelse{\equal{#2}{\true}}{#2}{m\mleft(#2\mright)}}}}
\newcommand*{\wl}[2][]{\wlfull{e_l, w_l#1, z_l#1}{#2}}
\newcommand*{\oiso}[1][m]{\sim_{#1}}
\newcommand*{\eiso}{\sim_{e}}
\newcommand*{\bisim}[1][m]{\sim_{#1}}

\newcommand*{\lb}{\mleft \lbrack}
\newcommand*{\rb}{\mright \rbrack}

\newcommand*{\norm}{\textsc{Norm}\xspace}
\newcommand*{\eq}{\textsc{Eq}\xspace}
\newcommand*{\bnd}{\textsc{Bnd}\xspace}
\newcommand*{\true}{\textsc{True}\xspace}

\usepackage{mleftright}
\usepackage{amsmath}
\usepackage{stmaryrd}
\usepackage[inline,shortlabels]{enumitem}
\usepackage{aligned-overset}
\usepackage{xspace}
\usepackage[normalem]{ulem}
\usepackage[mode=buildnew]{standalone}

\usepackage{ifthen}
\usepackage{aligned-overset}
\usepackage{acronym}

\acrodef{BAT}{basic action theory}
\acrodef{HTN}{hierarchical task network}
\acrodef{oi}{observationally indistinguishable}

\newboolean{techreport}
\setboolean{techreport}{true}

\ifthenelse{\boolean{techreport}}{
\AtEndDocument{\setcounter{TotPages}{8}}
}{}

\usepackage[createShortEnv,conf={restate, no link to proof}]{proof-at-the-end}

\newcommand{\remove}[1]{}

\newcommand*{\fullref}[1]{\hyperref[{#1}]{\autoref*{#1}~(\nameref*{#1})}}
\newcommand*{\subdefref}[2]{\hyperref[{#2}]{\autoref*{#1}.\ref*{#2}}}

\setcopyright{ifaamas}
\acmConference[AAMAS '23]{Proc.\@ of the 22nd International Conference
on Autonomous Agents and Multiagent Systems (AAMAS 2023)}{May 29 -- June 2, 2023}
{London, United Kingdom}{A.~Ricci, W.~Yeoh, N.~Agmon, B.~An (eds.)}
\copyrightyear{2023}
\acmYear{2023}
\acmDOI{}
\acmPrice{}
\acmISBN{}

\title{Abstracting Noisy Robot Programs}
\ifthenelse{\boolean{techreport}}{
\titlenote{This is an extended version of the original paper to be presented at AAMAS'23.}
}{
}

\hypersetup{
  pdftitle = {\@thetitle},
  pdfauthor = {Till Hofmann, Vaishak Belle}
}

\author{Till Hofmann}
\affiliation{
  \institution{RWTH Aachen University}
  \city{Aachen}
  \country{Germany}
}
\email{hofmann@kbsg.rwth-aachen.de}
\author{Vaishak Belle}
\affiliation{
  \institution{University of Edinburgh}
  \city{Edinburgh}
  \country{United Kingdom}
}
\email{vaishak@ed.ac.uk}


\begin{document}

\begin{abstract}
  Abstraction is a commonly used process to represent some low-level
system by a more coarse specification with the goal to omit unnecessary
details while preserving important aspects. While recent work on
abstraction in the situation calculus has focused on non-probabilistic
domains, we describe an approach to abstraction of probabilistic and
dynamic systems. Based on a variant of the situation calculus with
probabilistic belief, we define a notion of bisimulation that allows to
abstract a detailed probabilistic basic action theory with noisy
actuators and sensors by a possibly non-stochastic basic action theory.
By doing so, we obtain abstract Golog programs that omit unnecessary
details and which can be translated back to a detailed program for
actual execution. This simplifies the implementation of noisy robot
programs, opens up the possibility of using non-stochastic reasoning
methods (e.g., planning) on probabilistic problems, and provides domain
descriptions that are more easily understandable and explainable.

\end{abstract}

\pagestyle{fancy}
\fancyhead{}


\maketitle

\section{Introduction}

\emph{Abstraction} --- the ``process of mapping a representation of a problem onto a new representation'' \cite{giunchigliaTheoryAbstraction1992} --- is a ubiquitous concept both in human behavior and in computing systems, e.g.,
a simple activity such as buying milk involves dozens of actions that a human conveniently abstracts into a single task, and machine instructions (which itself are abstractions of physical processes) are abstracted by higher programming languages.
It has also seen widespread usage in several areas of artificial intelligence research \cite{saittaAbstractionArtificialIntelligence2013}, in particular in task planning.
Abstraction typically involves suppressing irrelevant information and therefore allows reasoning about complex problems that would otherwise be infeasible.
In the context of intelligent agents, abstraction typically serves three purposes \cite{belleAbstractingProbabilisticModels2020}:
\begin{enumerate*}[(1)]
  \item it provides a way to structure knowledge,
  \item it allows reasoning about larger problems by abstracting the problem domain, resulting in a smaller search space,
  \item it may provide more meaningful explanations and is therefore critical for \emph{explainable AI}.
\end{enumerate*}
The need for abstraction becomes particularly apparent when dealing with robotic systems: as a robot acts in a dynamic environment with imperfect sensors and actuators, its actions are inherently noisy.
As an example, a robot may intend to \move but may get stuck with some probability.
However, when programming such a robot, it is desirable to ignore those probabilistic aspects and instead work with a high-level and non-stochastic system, where the \move action always succeeds, for all of the reasons above:
Correctly designing a probabilistic domain is challenging, reasoning on such a domain is hard, and understanding how such a system operates is difficult.
This becomes even more important when considering a robot that may get a hardware upgrade: while the low-level behavior changes (e.g., a new sensor has a different noise profile), the high-level behavior should not be affected.
By using abstraction, we only need to update the low-level model and may keep the high-level program as is.

In this paper, we present an abstraction framework for robot programs with probabilistic belief:
Starting with the logic \ds \cite{belleReasoningProbabilitiesUnbounded2017}, a modal variant of the situation calculus with probabilistic belief, we describe a transition semantics for noisy \golog programs in Section~\ref{sec:dsg}.
Based on this transition semantics, we then propose a notion of abstraction of noisy programs, building on top of abstraction of probabilistic static models \cite{belleAbstractingProbabilisticModels2020} and non-stochastic dynamic models \cite{banihashemiAbstractionSituationCalculus2017}.
We do so by defining a notion of bisimulation of probabilistic dynamic systems in Section~\ref{sec:abstraction} and we show that the notions of sound and complete abstraction carry over.
We also demonstrate how this abstraction framework can be used to define a high-level domain, where noisy actions are abstracted away and thus, no probabilistic reasoning is necessary.
We conclude in Section~\ref{sec:conclusion}.

\section{Background and Related Work}
\paragraph{Reasoning about actions}
The situation calculus \cite{mccarthySituationsActionsCausal1963,reiterKnowledgeActionLogical2001} is a logical formalism for reasoning about dynamical domains based on first-order logic.
In the situation calculus, world states are represened explicitly as first-order terms called \emph{situations}, where \emph{fluents} describe (possibly changing) properties of the world and actions are axiomatized in \acfp{BAT}.
\golog~\cite{levesqueGOLOGLogicProgramming1997,degiacomoConGologConcurrentProgramming2000} is a programming language based on the situation calculus that allows to control the high-level behavior of robots.
\es \cite{lakemeyerSemanticCharacterizationUseful2011} is a modal variant and epistemic extension of the situation calculus, where situations are part of the semantics but do not appear as terms in the language.
\esg extends \es with a transition semantics for \golog programs, which has been used for program verification~\cite{classenPlanningVerificationAgent2013}.
The situation calculus, \es, and \esg are all deterministic and non-stochastic, i.e., the execution of an action always results in a unique successor state.
\textcite{degiacomoNondeterministicSituationCalculus2021} extend the situation calculus with non-deterministic actions, where the environment chooses one of several possible outcomes of an action.
\textcite{bacchusReasoningNoisySensors1999} extend the classical situation calculus with degrees of belief and noisy actions.
In a similar fashion, \ds \cite{belleReasoningProbabilitiesUnbounded2017} extends \es with degrees of belief and probabilistic actions, where the environment again may choose an outcome (possibly from an unbounded domain) with some pre-defined probability, allowing probabilistic representations of robot actions, in particular noisy sensors and actuators.
More recently, reasoning about actions in \ds has been shown to be amenable to regression~\cite{liuReasoningBeliefsMetabeliefs2021} and progression~\cite{liuProgressionBelief2021} analogous to regression and progression in \es and the classical situation calculus~\cite{reiterKnowledgeActionLogical2001}.

\paragraph{Abstraction}
\textcite{giunchigliaTheoryAbstraction1992} define abstraction generally as a mapping between a ground and an abstract formal system, such that the abstract representation preserves desirable properties while omitting unnecessary details to make it simpler to handle.
Abstraction has been widely used in several fields of AI \cite{saittaAbstractionArtificialIntelligence2013}.
\Ac{HTN} planning systems such as \textsc{SHOP2} \cite{nauSHOP2HTNPlanning2003} decompose tasks into subtasks to accomplish some overall objective, which has also been used in the situation calculus \cite{gabaldonProgrammingHierarchicalTask2002}.
Macro planners such as \textsc{MacroFF}~\cite{boteaMacroFFImprovingAI2005} combine action sequences into macro operators to improve planner performance, e.g., by collecting action traces from plan executions on robots \cite{hofmannInitialResultsGenerating2017}, or by learning them from training problems \cite{chrpaMUMTechniqueMaximising2014}.
Similarly, \textcite{saribaturOmissionbasedAbstractionAnswer2021} use abstraction in Answer Set Programming to reduce the search space, improving solver performance.
\textcite{cuiUniformAbstractionFramework2021} leverage abstraction for generalized planning, i.e., for finding general solutions for a set of similar planning problems.
Abstraction has also been used to analyze causal models \cite{rubensteinCausalConsistencyStructural2017,banihashemiActionsProgramsAbstract2022}. Of particular interest for this work is the notion of \emph{constructive abstraction} \cite{beckersAbstractingCausalModels2019}, where the refinement mapping partitions the low-level variables such that each cell has a unique corresponding high-level variable.
\textcite{holtzenSoundAbstractionDecomposition2018} describe an abstraction framework for probabilistic programs and also describe an algorithm to  generate abstractions.
%
\textsc{REBA}~\cite{sridharanREBARefinementbasedArchitecture2019} is a framework for robot planning that uses abstract and determistic ASP programs to determine a course of action, which are then translated to POMDPs for execution.
\textcite{banihashemiAbstractionSituationCalculus2017} describe a general abstraction framework based on the situation calculus, where a refinement mapping maps a high-level \ac{BAT} to a low-level \ac{BAT} and which is capable of online execution with sensing actions \cite{banihashemiAbstractionAgentsExecuting2018}.
The framework has been used to effectively synthesize plan process controllers in a smart factory scenario~\cite{degiacomoSituationCalculusController2022}.
In contrast to this work, they assume non-probabilistic and deterministic actions.
On the other hand, \textcite{belleAbstractingProbabilisticModels2020} defines abstraction in a probabilistic but static propositional language and describes a search algorithm to derive such abstractions.
In this paper, we build on the two approaches to obtain abstraction in a probabilistic and dynamic first-order language with an unbounded domain.

\section{The Logic \dsg}\label{sec:dsg}

We start by introducing the logic \dsg, which we will then use to define abstraction over noisy programs in \autoref{sec:abstraction}.
\dsg extends \ds~\cite{belleReasoningProbabilitiesUnbounded2017} with a transition semantics for \golog, analogous to how \esg~\cite{classenLogicNonterminatingGolog2008} extends \es~\cite{lakemeyerSemanticCharacterizationUseful2011}.
In the same way as \ds, the logic uses a countably infinite set of \emph{rigid designators} \rigid,  which allows to define quantification substitutionally.
Similar to \ds, \es, and \esg, it uses a possible-worlds semantics, where a world defines the state of the world not only initially but after any sequence of actions.
It uses the modal operator $[\cdot]$ to refer to the state after executing some program, e.g., $[ \delta ] \alpha$ states that $\alpha$ is true after every possible execution of the program $\delta$.
Additionally, it uses the modal operator $\belconnector$ to describe the agent's \emph{belief}, e.g., $\bel{\loc(2)}{0.5}$ states the the agent believes with degree $0.5$ to be in location $2$.

\subsection{Syntax}
\begin{definition}[Symbols of \dsg{}]
The symbols of the language are from the following vocabulary:
\begin{enumerate}
  \item infinitely many variables $x, y, \ldots, u, v, \ldots, a, a_1, \ldots$;
  \item rigid function symbols of every arity, e.g., $\mi{near}$, $\goto(x, y)$;
  \item fluent predicates of every arity, such as $\mi{\at(l)}$; we assume that this list contains the following distinguished predicates:
    \begin{itemize}
      \item $\poss$ to denote the executability of an action;
      \item $\oi$ to denote that two actions are indistinguishable from the agent's viewpoint; and
      \item $l$ that takes an action as its first argument and the action's likelihood as its second argument;
    \end{itemize}
  \item connectives and other symbols: $=$, $\wedge$, $\neg$, $\forall$, $\square$, $[\cdot]$,
    $\belconnector$.
\end{enumerate}
\end{definition}

\begin{definition}[Terms of \dsg{}]
  The set of terms of \dsg{} is the least set such that
  \begin{enumerate*}[(1)]
    \item every variable is a term,
    \item if $t_1, \ldots, t_k$ are terms and $f$ is a $k$-ary function symbol, then $f(t_1, \ldots, t_k)$ is a term.
  \end{enumerate*}
\end{definition}

As in \ds, we let $\rigid$ denote the set of all ground rigid terms and we assume that they contain the rational numbers, i.e., $\mathbb{Q} \subseteq \rigid$.

\begin{definition}[Formulas] \label{def:formulas}
  The \emph{formulas of \dsg{}} are the least set such that
  \begin{enumerate}
    \item if $t_1,\ldots,t_k$ are terms and $P$ is a $k$-ary predicate symbol,
      then $P(t_1,\ldots,t_k)$ is a formula,
    \item if $t_1$ and $t_2$ are terms,
      then $(t_1 = t_2)$ is a formula,
    \item if $\alpha$ and $\beta$ are formulas,
      $x$ is a variable, $\delta$ is a program (defined below),%
      \footnote{
        Note that although the definitions of formulas (\autoref{def:formulas}) and programs (\autoref{def:programs}) mutually depend on each other, they are still well-defined:
        programs only allow static situation formulas and static situation formulas may not refer to programs.
        Technically, we would first need to define static situation formulas, then programs, and then all formulas.
        For the sake of presentation, we omit this separation.
      }
      and
      $r \in \mathbb{Q}$,
      then $\alpha \wedge \beta$, $\neg \alpha$, $\forall x.\, \alpha$,
      $\square\alpha$, $[\delta]\alpha$,
      and $\bel{\alpha}{r}$ are formulas.
  \end{enumerate}
\end{definition}
We read $\square \alpha$ as ``$\alpha$ holds after executing any sequence of
actions'', $[\delta] \alpha$ as ``$\alpha$ holds after the execution of program $\delta$''
and $\bel{\alpha}{r}$ as ``$\alpha$ is believed with probability $r$''.%
\footnote{The original version of the logic also has an only-knowing modal operator \obelconnector, which captures the idea that something and only that thing is known. For the sake of simplicity, we ignore this operator in our presentation.}
We also write $\know{\alpha}$ for $\bel{\alpha}{1}$, to be read as ``$\alpha$ is known''.%
\footnote{We use ``knowledge'' and ``belief'' interchangeably, but do not require that knowledge be true in the real world (i.e., weak S5).}
We use \true as abbreviation for $\forall x \left(x = x\right)$ to denote truth. 
For a formula $\alpha$, we write $\alpha^x_r$ for the formula resulting from $\alpha$ by substituting every occurrence of $x$ with $r$.
For a finite set of formulas $\Sigma = \{ \alpha_1, \ldots, \alpha_n \}$, we may just write $\Sigma$ for the conjunction $\alpha_1 \wedge \ldots \wedge \alpha_n$, e.g., $\know{\Sigma}$ for $\know{(\alpha_1 \wedge \cdots \wedge \alpha_n)}$.
A predicate symbol with terms from $\rigid$ as arguments is called a \emph{primitive formula}, and we denote the set of primitive formulas with $\atoms$.
Furthermore, a formula is called \emph{bounded} if it contains no $\square$ operator,
\emph{static} if it contains no $[\cdot]$ or $\square$ operators,
\emph{objective} if it contains no \belconnector or \know{},
and
{\em fluent} if it is static and does not mention \poss{}, \belconnector, or \know{}.

\newcommand*{\smid}{\ensuremath{\:\mid\:}}

Finally, we define the syntax of \golog{} programs referred to by the operator $[\delta]$.

\begin{definition}[Programs] \label{def:programs}
  \[
    \delta ::= t \smid \alpha? \smid \delta_1 ; \delta_2 \smid \delta_1|\delta_2
    \smid \pi x.\, \delta
    \smid \delta^*
  \]
  where $t$ is a ground rigid term and $\alpha$ is a static formula. A
  program consists of actions $t$, tests $\alpha?$, sequences
  $\delta_1;\delta_2$, nondeterministic branching $\delta_1 | \delta_2$,
  nondeterministic choice of argument $\pi x.\, \delta$,
  and nondeterministic iteration $\delta^*$.
\end{definition}

Note that we do not allow interleaved concurrency $\delta_1 \| \delta_2$ known from \congolog~\cite{degiacomoConGologConcurrentProgramming2000}.%
\footnote{The reason will become apparent later on. Intuitively, if we allow interleaved concurrency, then the low-level program could pause the execution of a high-level action and continue with a different high-level action, possibly leading to different effects.
  This significantly complicates the formal treatment relating the probabilities of high-level worlds to their low-level counterparts.
}
We also use $\mi{nil}$ as abbreviation for $\true?$, the empty program that always succeeds.
Similarly to formulas, $\delta^x_r$ denotes the program resulting from $\delta$ by substituting every $x$ with $r$.
Furthermore, we define $\gif \ldots \gfi$ and $\gwhile \ldots \gdone$ as syntactic sugar as follows:
\begin{align*}
  \gif \phi \gthen \delta_1 \gelse \delta_2 \gfi
  &\eqdef
  (\phi?; \delta_1) \mid (\neg \phi?; \delta_2)
  \\
  \gif \phi_1 \gthen \delta_1 \gelif \phi_2 \gthen \delta_2 \gfi
  &\eqdef
  (\phi_1?; \delta_1) \mid (\neg \phi_1 \wedge \phi_2?; \delta_2)
  \\
  \gwhile \phi \gdo \delta \gdone
  &\eqdef
  (\phi?; \delta)^*; \neg \phi?
\end{align*}

\subsection{Semantics}

As described above, the operator \belconnector describes the degree of belief.
In order to capture noisy actions and sensors, we need to talk about the \emph{likelihood of possible outcomes} as well as the fact that when a noisy action is executed, the intended outcome may not be the same as the desired outcome.
The latter is captured using the notion of \emph{observational indistinguishability}.
Both likelihood of possible outcomes and observational indistinguishability are built into the worlds using distinguished symbols and then modelled using \emph{basic action theories}, as described in \autoref{sec:bat}.

Similar to \ds, the semantics of \dsg is given in terms of \emph{possible worlds}, where
a world defines the truth of each fluent both initially and after any sequence of actions, also called traces:
\begin{definition}[Trace]
  A trace $z = \la a_1, \ldots, a_n \ra$ is a finite sequence of \rigid.
  We denote the set of traces as \traces and the empty trace with $\la\ra$.
\end{definition}
A world defines the truth of each ground atom from $\atoms$ not only initially but after any sequence of actions:
\begin{definition}[World]
  A world is mapping $w : \atoms \times \traces \rightarrow \{0, 1\}$.
  The set of all worlds is denoted as \worlds.
\end{definition}

We require that every world $w \in \worlds$ defines a unary predicate $\poss$, a binary predicate $l$ that behaves like a function (i.e., there is exactly one $q \in \mathbb{Q}$ such that $w[l(a, q), z] = 1$ for any $a, z$),
as well as an equivalence relation $\mi{oi} \subseteq \rigid \times \rigid$, which define the possibility, the likelihood, and the observational indistinguishability of actions.

We call a pair $\left(w, z\right) \in \worlds \times \traces$ a \emph{state}, we denote the set of all states with \states, and we use $\stateset, \stateset_i, \ldots \subseteq \states$ to denote sets of states.

Given a state $(w, z)$, the predicate $l(a, q)$ states that the action likelihood of action $a$ in state $(w, z)$ is equal to $q$.
We can inductively apply $l$ to compute the likelihood of a sequence:
\begin{definition}[Action Sequence Likelihood]
  The action sequence likelihood $l^*: \worlds \times \mathcal{Z} \rightarrow \mathbb{Q}^{\geq 0}$ is defined inductively:
  \begin{itemize}
    \item $l^*\mleft( w, \la\ra \mright) = 1$ for every $w \in \worlds$,
    \item $l^*\mleft(w, z \cdot r \mright) = l^*\mleft(w, z\mright) \times q$ where $w \mleft[ l(r, q), z \mright] = 1$.
  \end{itemize}
\end{definition}
Next, to deal with partially observable states, we define:
\begin{definition}[Observational indistinguishability]
  ~
  \begin{enumerate}[left=0pt]
    \item
      Given a world $w \in \worlds$, we define the relation $\oisim \subset \mathcal{Z} \times \mathcal{Z}$ inductively:
      \begin{itemize}[labelindent=0pt]
        \item $\la\ra \oisim z'$ iff $z' = \la\ra$
        \item $z \cdot r \oisim z'$ iff $z' = z^* \cdot r^*$, $z \oisim z^*$, and $w \mleft[ \oi(r,r^*), z \mright] = 1$
      \end{itemize}
    \item
      We say $w$ is \acfi{oi} from $w'$, written $w \oicomp w'$ iff for all $a, a' \in \rigid$, $z \in \traces$: $w \mleft[ \mi{oi}(a,a'),z \mright] = w' \mleft[ \mi{oi}(a,a'),z \mright]$.
    \item
      For $w, w' \in \worlds$, $z, z' \in \mathcal{Z}$, we say $\left(w, z\right)$ is \ac{oi} from $\left(w', z'\right)$,  written $\left(w, z\right) \oicomp \left(w', z'\right)$, iff $w \oicomp w'$ and $z \oisim z'$.
  \end{enumerate}
\end{definition}
Intuitively, $z \oisim z'$ means that the agent cannot distinguish whether it executed $z$ or $z'$.
For states, $\left(w, z\right) \oicomp \left(w', z'\right)$ is to be understood as ``if the agent believes to be in state $\left(w, z\right)$, it may also actually be in state $\left(w', z'\right)$'', i.e., it cannot distinguish the possible worlds $w, w'$ and traces $z, z'$.
As \oicomp is an equivalence relation, the set of its equivalence classes on a set of states \stateset induces a partition, which we denote with $\stateset / \oicomp$.

As another notational device, we extend the executability of an action to traces:
\begin{definition}[Executable trace]
  For a trace $z$, we define $\exec(z)$ inductively:
  \begin{itemize}
    \item for $z = \la\ra$, $\exec(z) \eqdef \true$
    \item for $z = a \cdot z'$, $\exec(z) \eqdef \poss(a) \wedge [a] \exec(z')$
  \end{itemize}
\end{definition}

As in \bhl and \ds, it is possible to permit the agent to entertain any set of initial distributions.
As an example, the initial theory could say that $\bel{p}{0.5} \vee \bel{p}{0.6}$, which says that the agent is not sure about the distribution of $p$.
In this case, there would be two distributions in the epistemic state $e$.
As another example, if we say $\bel{p \vee q}{1}$, then this says that the disjunction is believed with probability $1$, but it does not specify the probability of $p$ or $q$, resulting in infinitely many distributions that are compatible with this constraint.
Thus, not committing to a single distribution results in higher expressivity in the representation of uncertainty.

\begin{definition}[Compatible States]
  Given an epistemic state $e$, a world $w$, a trace $z$, and a formula $\alpha$, we define the states $\stateset^{e,w,z}_\alpha$ compatible to $\left(e, w, z\right)$ wrt to $\alpha$:
  \[
    \stateset^{e,w,z}_\alpha = \{ (w', z') \mid (w', z') \oicomp (w, z), e, w' \models \exec(z') \wedge [z']\alpha \}
  \]
\end{definition}
We may write $\stateset_\alpha$ for $\stateset^{e,w,z}_\alpha$ if $e,w,z$ are clear from the context.

To define the semantics of belief, we first define \emph{epistemic states}, which assign probabilities to worlds:
\begin{definition}[Epistemic state]
  A distribution is a mapping $\worlds \rightarrow \mathbb{R}^{\geq 0}$.
  An epistemic state is any set of distributions.
\end{definition}
We need to be able to sum over uncountably many worlds, which we do as follows:
\begin{definition}[Normalization]
  ~ \\
  For any distribution $d$ and any set $\mathcal{V} = \left\{\left(w_1, z_1\right), \left(w_2, z_2\right), \ldots \right\}$, we define:
  \begin{enumerate}
    \item $\bnd(d, \mathcal{V}, r)$ iff there is no $k$ such that
      \[
        \sum_{i=1}^k d(w_i) \times l^*(w_i, z_i) > r
      \]
    \item $\eq\mleft(d, \mathcal{V}, r\mright)$ iff $\bnd\mleft(d, \mathcal{V}, r\mright)$ and there is no $r' < r$ such that $\bnd\mleft(d, \mathcal{V}, r'\mright)$ holds.
    \item For any $\mathcal{U} \subseteq \mathcal{V}$: $\norm\mleft(d, \mathcal{U}, \mathcal{V}, r\mright)$ iff $\exists b \neq 0$ such that $\eq\mleft(d, \mathcal{U}, b \times r\mright)$ and $\eq\mleft(d, \mathcal{V}, b\mright)$.
  \end{enumerate}
\end{definition}
Intuitively, given $\norm(d, \mathcal{V}, r)$, $r$ can be seen as the normalization of the weights of worlds in $\mathcal{V}$ in relation to the set of all worlds $\mathcal{W}$ as accorded by $d$.
The conditions \bnd and \eq are auxiliary conditions to define \norm, where $\bnd(d, \mathcal{V}, r)$ states that the weight of worlds in $\mathcal{V}$ is bounded by $b$ and $\eq(d, \mathcal{V}, r)$ expresses that the weight of worlds in $\mathcal{V}$ is equal to $b$.
\textcite{belleFirstorderLogicProbability2016} have shown that although the set of worlds $\mathcal{W}$ is in general uncountable, this leads to a well-defined summation over the weights of worlds.

To simplify notation,
we also write $\norm(d, \mathcal{U}, \mathcal{V}) = r$ for $\norm(d, \mathcal{U}, \mathcal{V}, r)$.
Furthermore, we write $\norm(d_1, \mathcal{U}_1, \mathcal{V}_1) = \norm(d_2, \mathcal{U}_2, \mathcal{V}_2)$ if there is an $r$ such that
$\norm(d_1, \mathcal{U}_1, \mathcal{V}_1, r)$ and $\norm(d_2, \mathcal{U}_2, \mathcal{V}_2, r)$.
Finally, we write
\[
  \norm(d, \mathcal{U}_1, \mathcal{V}) + \norm(d, \mathcal{U}_2, \mathcal{V}) = r
\]
if
$\norm(d, \mathcal{U}_1, \mathcal{V}, r_1)$,
$\norm(d, \mathcal{U}_2, \mathcal{V}, r_2)$, and $r = r_1 + r_2$.

We continue with the program transition semantics, which defines the traces resulting from executing some program $\delta$.
The transition semantics is defined in terms of \emph{configurations} $\la z, \rho \ra$, where $z$ is a trace describing the actions executed so far and $\rho$ is the remaining program.
In some places, the transition semantics refers to the truth of formulas (see \autoref{def:truth} below).%
\footnote{As above, although they depend on each other, the semantics is well-defined, as the transition semantics only refers to static formulas which may not contain programs.}%
\newcommand{\warrow}{\ensuremath{\overset{e,w}{\longrightarrow}}}
\newcommand*{\final}{\ensuremath{\mathcal{F}^{e,w}}}
\begin{definition}[Program Transition Semantics]\label{def:trans}
  The transition relation \warrow{} among configurations, given
  an epistemic state $e$ and a world $w$, is the least set satisfying
  \begin{enumerate}
    \item $\la z, a\ra \warrow \la z \cdot a, \mi{nil} \ra$ if $w, z \models \poss(a)$
    \item $\la z, \delta_1;\delta_2 \ra \warrow
      \la z \cdot a, \gamma;\delta_2 \ra$,
      if $\la z,\delta_1 \ra \warrow \la z \cdot a, \gamma \ra$,
    \item $\la z, \delta_1;\delta_2 \ra \warrow \la z \cdot a, \delta' \ra$
      if \\ $\la z, \delta_1 \ra \in \final$ and
      $\la z, \delta_2 \ra \warrow \la z \cdot a, \delta' \ra$
    \item $\la z, \delta_1 | \delta_2 \ra \warrow \la z \cdot a, \delta' \ra$
      if \\ $\la z, \delta_1 \ra \warrow \la z \cdot a, \delta' \ra$
      or $\la z, \delta_2 \ra \warrow \la z \cdot a, \delta' \ra$
    \item $\la z, \pi x.\, \delta \ra \warrow \la z \cdot a, \delta' \ra$,
      if \\ $\la z, \delta^x_r \ra \warrow \la z \cdot a, \delta' \ra$ for some
      $r \in \rigid$
    \item $\la z, \delta^* \ra \warrow \la z \cdot a, \gamma; \delta^* \ra$ if
      $\la z, \delta \ra \warrow \la z \cdot a, \gamma \ra$
  \end{enumerate}
  The set of final configurations \final{} is the smallest set such that
  \begin{enumerate}
    \item $\la z, \alpha? \ra \in \final$ if $e, w, z \models \alpha$,
    \item $\la z, \delta_1;\delta_2 \ra \in \final$
      if $\la z, \delta_1 \ra \in \final$ and $\la z, \delta_2 \ra \in \final$
    \item $\la z, \delta_1 | \delta_2 \ra \in \final$
      if $\la z, \delta_1 \ra \in \final$,
      or $\la z, \delta_2 \ra \in \final$
    \item $\la z, \pi x.\,\delta \ra \in \final$
      if $\la z, \delta^x_r \ra \in \final$ for some $r \in \rigid$
    \item $\la z, \delta^* \ra \in \final$
  \end{enumerate}
\end{definition}
We also write $\warrow^*$ for the transitive closure of $\warrow$.
For a primitive action $a$, the interpreter may take a transition if $a$ is currently possible.
For a sequence of sub-programs $\delta = \delta_1; \delta_2$, the interpreter may take a transition following $\delta_1$, or it may take a transition following $\delta_2$ if $\delta_1$ is final in the current configuration.
In the case of nondeterministic branching $\delta_1 \vert \delta_2$, it may follow the transitions of the first or the second sub-program.
For the nondeterministic pick operator $\pi x.\, \delta$, it may follow any transition that results from the program $\delta^x_r$, where $x$ is substituted by some ground term $r$.
Finally, for nondeterministic iteration $\delta^*$, the interpreter may take the same transitions as $\delta$ (i.e., continue with another iteration).

For the final configurations, atomic tests $\alpha?$ are final if $\alpha$ is satisfied in the current configuration.
The sequence of sub-programs $\delta_1; \delta_2$ is final if both sub-programs are final.
For nondeterministic branching, the program $\delta_1 \vert \delta_2$ is final if either sub-program is final.
Similarly, for $\pi x.\, \delta$, the program is final if it is final for any substitution of $x$.
Nondeterministic iteration $\delta^*$ is final, i.e., the interpreter may always decide to stop (and not continue with the next iteration).

Following the transition semantics for a given program $\delta$, we obtain a set of \emph{program traces}:
\begin{definition}[Program Traces]
  ~\\
  Given an epistemic state $e$, a world $w$, and a trace $z$, the set $\|\delta\|^z_{e,w}$ of traces of program $\delta$ is defined as the following set:
  \begin{multline*}
    \|\delta\|^z_{e, w} =
    \\
    \{ z' \in \mathcal{Z} \mid
      \la z, \delta \ra \warrow^* \la z \cdot z', \delta' \ra
    \text{ and } \la z \cdot z', \delta' \ra \in \final \}
  \end{multline*}
\end{definition}

Compared to \esg, this transition semantics also refers to the epistemic state $e$, as test formulas can also mention belief operators.
Additionally, in contrast to \esg, it only allows a transition for an atomic action if the action is possible in the current state.
Also, while \esg allows infinite traces, we only allow finite traces, as we do not include temporal formulas in the logic.

Finally, 
we can define the semantics for \dsg formulas:
\begin{definition}[Truth of Formulas]\label{def:truth}
  Given an epistemic state $e$, a world $w$, and a formula $\alpha$,
  we define for every $z \in \mathcal{Z}$:
  \begin{enumerate}
    \item $e, w, z \models F\mleft(t_1, \ldots, t_k\mright)$ iff $w \mleft\lbrack F\mleft(t_1, \ldots, t_k\mright), z \mright\rbrack = 1$
    \item $e, w, z \models \bel{\alpha}{r}$
      iff $\forall d \in e:\: \norm\mleft(d, \stateset_\alpha, \stateset_\true, r\mright)$
    \item $e, w, z \models \left(t_1 = t_2\right)$ iff $t_1$ and $t_2$ are identical
    \item $e, w, z \models \alpha \wedge \beta$ iff $e, w, z \models \alpha$ and
      $e, w, z \models \beta$
    \item $e, w, z \models \neg \alpha$ iff $e, w, z \not\models \alpha$
    \item $e, w, z \models \forall x.\, \alpha$ iff $e, w, z \models \alpha^x_r$ for
      all $r \in \rigid$.
    \item $e, w, z \models \square \alpha$ iff $e, w, z \cdot z' \models \alpha$
      for all $z' \in \mathcal{Z}$
    \item $e, w, z \models [\delta]\alpha$ iff
     $e, w, z \cdot z' \models \alpha$
     for all $z' \in \|\delta\|^z_{e,w}$.
     \label{def:truth:program}
  \end{enumerate}
\end{definition}

Note in particular that Item 2 states that the degree of belief in a formula is obtained by looking at the normalized weight of the possible worlds that satisfy the formula.

We write $e, w \models \alpha$ for $e, w, \la\ra \models \alpha$.
Also, if $\alpha$ is objective, we write $w, z \models \alpha$ for $e, w, z \models \alpha$ and $w \models \alpha$ for $w, \la\ra \models \alpha$.
Additionally, for a set of sentences $\Sigma$, we write $e, w, z \models \Sigma$ if $e, w, z \models \phi$ for all $\phi \in \Sigma$, and $\Sigma \models \alpha$ if $e, w \models \Sigma$ entails $e, w \models \alpha$ for every model $\left(e, w\right)$.

\subsection{Basic Action Theories}\label{sec:bat}
A \acf{BAT} defines the effects of all actions of the domain, as well as the initial state:
\begin{definition}[Basic Action Theory]
  Given a finite set of predicates $\mathcal{F}$ including \oi and $l$, a set $\Sigma$
  of sentences is called a \acf{BAT} over $\mathcal{F}$ iff
  $\Sigma = \Sigma_0 \cup \Sigma_\text{pre} \cup \Sigma_\text{post}$, where
  $\Sigma$ mentions only fluent predicates in $\mathcal{F}$ and
  \begin{enumerate}
    \item $\Sigma_0$ is any set of fluent sentences,
    \item $\Sigma_\text{pre}$ consists of a single sentence
      of the form $\square
      \poss(a) \equivspace \pi$, where $\pi$ is a fluent formula with free variable $a$,%
      \footnote{We assume that free variables are universally quantified from the outside, $\square$ has lower syntactic precedence than the logical connectives, and $[\cdot]$ has the highest priority, so that $\square \poss(a) \equiv \pi$ stands for $\forall a.\, \square \left(\poss(a) \equiv \gamma\right)$ and $\square \poss(a) \supset \left([a]F(\vec{x}) \equiv \gamma_F\right)$ stands for $\forall a,\vec{x}.\, \square \left(\poss(a) \supset \left([a]F(\vec{x}) \equiv \gamma_F\right)\right)$.}
    \item $\Sigma_\text{post}$ is a set of sentences, one for each fluent predicate $F \in \mathcal{F}$, of the form $\square \poss(a) \supset \left([a]F(\vec{x}) \equivspace \gamma_F\right)$, and where $\gamma_F$ is a fluent formula with free variables among $a$ and $\vec{x}$.
  \end{enumerate}
\end{definition}
Given a \ac{BAT} $\Sigma$, we say that a program $\delta$ is a program over $\Sigma$ if it only mentions fluents and actions from $\Sigma$.

Note that the successor state axioms slightly differ from \es and \esg, where they have the form $\square [a]F(\vec{x}) \equivspace \gamma_F$.
In contrast to \es and \esg, the successor state axioms in \dsg \acp{BAT} only define the effects of an action if the action is currently possible and otherwise do not make any statement about the action effects.
This is necessary because we include $\poss(a)$ in the transition semantics (\autoref{def:trans}).
To understand why, consider the following example: if $w, z \models \neg \poss(a)$, then by \subdefref{def:truth}{def:truth:program}, $w, z \models [a] \neg F$ is vacuously true for any $0$-ary fluent $F$ because there is no trace $z' \in \|a\|^z_{e,w}$.
This would be contradicting to a successor state axiom $\square [a] F \equiv \gamma_F$.
Restricting the successor state axiom to possible actions avoids this issue.%
\footnote{
  \textcite{classenPlanningVerificationAgent2013} proposes a different solution by allowing an action transition even if the action is impossible and then augmenting the program by guarding each action with a test $\poss(a)?$.
  We prefer the presented solution because the transition semantics only allows actions that are actually possible without augmenting the program.
}

\subsubsection{A Noisy Basic Action Theory}
We present a \ac{BAT} for a simple robotics scenario with noisy actions, inspired from \cite{bacchusReasoningNoisySensors1999,belleReasoningProbabilitiesUnbounded2017}.
In this scenario, a robot moves towards a wall and it is equipped with a sonar sensor that can measure the distance to the wall.
A \ac{BAT} \sigmal defining this scenario may look as follows:
\begin{itemize}[left=0pt .. \parindent]
  \item
    A \move action is possible if the robot moves either one step to the back or to the front.
    A \sonar action is always possible:
    \begin{align*}
      \square \poss(a) \equivspace
      & \exists x,y \left(a = \move(x,y) \wedge \left(x = 1 \vee x = -1\right)\right)
   \\ &
      \vee \exists z \left(a = \sonar(z)\right)
    \end{align*}
  \item After doing action $a$, the robot is at position $x$ if $a$ is a \move action that moves the robot to location $x$, if $a$ is a \sonar action that measures distance $x$, or if $a$ is neither of the two actions and the robot was at location $x$ before%
    \begin{align*}
      \square \poss(&a) \supset \big(\lb a \rb \loc(x) \equivspace \\
      &\exists y,z, \left(a = \move(y,z) \wedge \loc(l) \wedge x = l + z \right)
      \\
      & \vee a = \sonar(x)
      \\
      & \vee \neg \exists y, z \left(a = \move(y,z) \vee a = \sonar(y)\right) \wedge \loc(x)\big)
    \end{align*}
  \item For the \sonar action, the likelihood that the robot measures the correct distance is $0.8$, the likelihood that it measures a distance with an error of $\pm 1$ is $0.1$.
    Furthermore, for the \move action, the likelihood that the robot moves the intended distance $x$ is $0.6$, the likelihood that the actual movement $y$ is off by $\pm 1$ is $0.2$:
    \begin{align*}
      \square &l(a, u) \equivspace \\
      & \exists z \left( a = \sonar(z) \wedge \loc(x) \wedge u = \Theta(x, z, .8, .1)\right)
      \\
      &  \vee \exists x,y \left(a = \move(x, y) \wedge u = \Theta(x, y, .6, .2)\right)
      \\
      &  \vee \neg \exists x,y,z \left(a = \move(x,y) \vee a = \sonar(z)\right) \wedge u = .0
    \end{align*}
    where $\Theta(u,v,c,d) =
      \begin{cases} c & \text{ if } u = v \\ d & \text{ if } |u-v| = 1 \\ 0 & \text{ otherwise} \end{cases}$.
  \item  The robot cannot detect the distance that it has actually moved, i.e., any two actions $\move(x, y)$ and $\move(x, z)$ are o.i.:
    \begin{align*}
      \square \oi(a, a') \equivspace &a = a' \vee
      \\
            & \exists x, y, z \left(a = \move(x, y) \wedge a' = \move(x, z)\right)
    \end{align*}
  \item Initially, the robot is $\SI{3}{\metre}$ away from the wall: $\loc(x) \equivspace x = 3$
  \end{itemize}

Based on this \ac{BAT}, we define a program that first moves the robot close to the wall and then back:\footnote{The unary $\move(x)$ can be understood as abbreviation $\move(x) \eqdef \pi y\, \move(x, y)$, where nature nondeterministically picks the distance $y$ that the robot really moved (similarly for $\sonar()$).}
\begin{align*}
   &\sonar();
   \\
  & \gwhile \neg \know{\exists x \left(\loc(x) \wedge x \leq 2\right)}
  \gdo \move(-1); \sonar() \gdone;
  \\ &\gwhile \neg \know{\exists x \left(\loc(x) \wedge x > 5\right)}
  \gdo \move(1); \sonar() \gdone
\end{align*}

The robot first measures its distance to the wall and then moves closer until it knows that its distance to the wall is less than $\SI{2}{\metre}$.
Afterwards, it moves away until it knows that is more than $\SI{5}{\metre}$ away from the wall.
As the robot's \move action is noisy, each \move is followed by \sonar to measure how far it is away from the wall.
One possible execution trace of this program may look as follows:
\begin{equation}\label{eqn:low-level-trace}
\begin{aligned}
  z_l = \langle &\sonar(3), \move(-1, 0), \sonar(3), \move(-1, -1),
  \\ &
  \sonar(2), \move(-1, -1), \sonar(1), \move(1, 1),
   \\ &
  \sonar(3), \move(1, 1),
  \sonar(2), \move(1, 1),
  \\ &
  \sonar(4), \move(1, 1), \sonar(6)
  \rangle
\end{aligned}
\end{equation}
First, the robot (correctly) senses that it is $\SI{3}{\metre}$ away from the wall and starts moving.
However, the first \move does not have the desired effect: the robot intended to move by $\SI{1}{\metre}$ but actually did not move (indicated by the second argument being $0$).
After the second \move, the robot is  at $\loc(2)$, as it started at $\loc(3)$ and moved successfully once.
However, as its sensor is noisy and it measured $\sonar(2)$, it believes that it could also be at $\loc(3)$.
For safe measure, it executes another \move and then senses $\sonar(1)$, after which it knows for sure that it is at a distance $\leq \SI{2}{\metre}$.
In the second part, the robot moves back until it knows that it has reached a distance $> \SI{5}{\metre}$.
As this simple example shows, the trace $z_l$ is already quite hard to understand.
While it is clear from the \ac{BAT} what each action does, the robot's intent is not immediately obvious and the trace is cluttered with noise and low-level details.

\subsubsection{An Abstract Basic Action Theory}
We present a second, more abstract \ac{BAT} for the same scenario but without noisy actions:
\begin{itemize}[left=0pt .. \parindent]
  \item The robot may do action $a$ if $a$ is a \goto action to a valid location:\footnote{For the sake of brevity, we only allow the robot to go to $\near$ or $\far$ and omit $\midpos$.}
    \[
      \square \poss(a) \equivspace
      a = \goto(\near) \vee a = \goto(\far)
    \]
  \item After doing action $a$, the robot is at location $l$ if $a$ is the action $\goto(l)$ or if $a$ is no \goto action and the robot has been at $l$ before:
    \begin{multline*}
      \square \poss(a) \supset
      \\
      \big(\lb a \rb \at(l) \equivspace
      a = \goto(l) \vee \neg \exists x \left(a = \goto(x)\right) \wedge \at(l)\big)
    \end{multline*}
  \item The action likelihood axiom states that no action is noisy: 
    \begin{align*}
      \square l(a, u) \equivspace &(a = \goto(\near) \vee a = \goto(\far)) \wedge u = 1.0
      \\
              &\vee \neg \exists x \left(a = \goto(x)\right) \wedge u = 0.0
    \end{align*}
  \item The agent can distinguish all actions: $\square \oi(a, a') \equivspace a = a'$
  \item Initially, the robot is in the middle: $\at(l) \equivspace l = \midpos$
\end{itemize}

In the remainder of this paper, we will connect the low-level \ac{BAT} \sigmal with the high-level \ac{BAT} \sigmah by using \emph{abstraction}.


\section{Abstraction}\label{sec:abstraction}
In this section, we define the abstraction of a low-level \ac{BAT} $\Sigma_l$ with a high-level \ac{BAT} $\Sigma_h$.
This will allow us to construct abstract \golog programs over the high-level \ac{BAT}, which are equivalent and can be translated to some program over the low-level \ac{BAT}.
For the sake of simplicity\footnote{The technical results do not hinge on this, but allowing arbitrary epistemic states would make the main results and proofs more tedious. For the general case, we need to set up for every distribution on the high level a corresponding distribution on the low level and establish a bisimulation for each of those pairs.}, we assume in the following that an epistemic state $e$ is always a singleton, i.e., $e_h = \{ d_h \}$ and $e_l = \{ d_l \}$.
To translate the high-level \ac{BAT} $\Sigma_h$ into the low-level \ac{BAT} $\Sigma_l$, we map $\Sigma_h$ to $\Sigma_l$ by mapping each high-level fluent to a low-level formula, and every high-level action to a low-level program:
\begin{definition}[Refinement Mapping]
  Given two basic action theories $\Sigma_l$ over $\mathcal{F}_l$ and $\Sigma_h$ over $\mathcal{F}_h$. The function $m$  is a \emph{refinement mapping} from $\Sigma_h$ to $\Sigma_l$ iff:
  \begin{enumerate}
    \item For every action $a(\vec{x})$ mentioned in $\Sigma_h$, $m\mleft(a\mleft(\vec{x}\mright)\mright) = \delta_a\mleft(\vec{x}\mright)$, where $\delta_a\mleft(\vec{x}\mright)$ is a Golog program over the low-level theory $\Sigma_l$ with free variables among $\vec{x}$.
    \item For every fluent predicate $F \in \mathcal{F}_h$, $m\mleft(F(\vec{x})\mright) = \phi_F\mleft(\vec{x}\mright)$, where $\phi_F\mleft(\vec{x}\mright)$ is a static formula over $\mathcal{F}_l$ with free variables among $\vec{x}$.
  \end{enumerate}
\end{definition}

For a formula $\alpha$ over $\mathcal{F}_h$, we also write $m(\alpha)$ for the formula obtained by applying $m$ to each fluent predicate and action mentioned in $\alpha$.
For a trace $z = \la a_1, a_2, \ldots \ra$ of actions from $\Sigma_h$, we also write $m(z)$ for $\la m(a_1), m(a_2), \ldots \ra$.
For a program $\delta$ over $\Sigma_h$, the program $m(\delta)$ is the same program as $\delta$ with each primitive action $a$ replaced by $m(a)$ and each formula $\alpha$ replaced by $m(\alpha)$.

Continuing our example, we define a refinement mapping that maps \sigmah to \sigmal by mapping each high-level fluent to a low-level formula and each high-level action to a low-level program:
\begin{itemize}[left=0pt .. \parindent]
  \item The high-level fluent $\at(l)$ is mapped to a low-level formula by translating the distance to the locations $\near$, $\midpos$, and $\far$:
    \begin{align*}
      \at(l) \mapsto\;
      &l = \near \wedge \exists x \left(\loc(x) \wedge x \leq 2\right)
    \\
      \vee &l = \midpos \wedge \exists x \left(\loc(x) \wedge x > 2 \wedge x \leq 5\right)
      \\
       \vee &l = \far \wedge \exists x \left(\loc(x) \wedge x > 5\right)
    \end{align*}
  \item The action $\goto$ is mapped to a program that guarantees that the robot reaches the right position:
    \begin{align*}
       &\goto(x) \mapsto
       \sonar();
       \\
       & \gif x = \near \gthen \\
       & \, \gwhile \neg \know{\exists x \left(\loc(x) \wedge x \leq 2\right)}
       \gdo \move(-1); \sonar() \gdone
       \\
       & \gelif x = \far \gthen \\
       & \, \gwhile \neg \know{\exists x \left(\loc(x) \wedge x > 5\right)}
       \gdo \move(1); \sonar() \gdone
       \gfi
    \end{align*}
\end{itemize}

To show that a high-level \ac{BAT} indeed abstracts a low-level \ac{BAT}, we first define a notion of isomorphism, intuitively stating that two states satisfy the same fluents:
\begin{definition}[Objective Isomorphism]\label{def:oiso}
  ~ \\
  We say $\left(w_h,z_h\right)$ is objectively $m$-isomorphic to $\left(w_l,z_l\right)$, written $\left(w_h,z_h\right) \oiso \left(w_l,z_l\right)$ iff
      for every atomic formula $\alpha$ mentioned in $\Sigma_h$:
      \[
        w_h, z_h \models \alpha \textrm{ iff } w_l, z_l \models m\mleft(\alpha\mright)
      \]
\end{definition}

Additionally, because we need to relate degrees of belief, we need to connect the two \acp{BAT} in terms of epistemic states.
To do so, we define epistemic isomorphism as follows:
\begin{definition}[Epistemic Isomorphism]\label{def:eiso}
  ~ \\
  For every $(w_h, z_h) \in \states$ and $\stateset_l \subseteq \states$,
  we say that $\left(d_h,w_h,z_h\right)$ is epistemically $m$-isomorphic to $\left(d_l, \stateset_l\right)$, written $\left(d_l,w_h,z_h\right) \eiso \left(d_l,\stateset_l\right)$ iff
  for the partition $P = \stateset_l / \oicomp$, for each $\stateset_l^i \in P$ and $\left(w_l^i, z_l^i\right) \in \stateset_l^i$:
    \begin{multline*}
      \norm(d_h, \{ \left(w_h, z_h\right) \}, \wh{\true})
      = \norm(d_l, \stateset_l^i, \wl[^i]{\true})
    \end{multline*}
\end{definition}

The intuition of epistemic isomorphism is as follows: As the high-level state $\left(w_h, z_h\right)$ is more abstract than the low-level state $\left(w_l, z_l\right)$, multiple low-level states may be isomorphic to the same high-level state.
Hence, each high-level state is mapped to a set of low-level states.
To be epistemically isomorphic, they must entail the same beliefs, so the corresponding normalized weights must be equal.
However, we do not require the low-level states $\stateset_l$ to be \ac{oi}.
Indeed, since we will have a high-level action corresponding to many low-level actions, low-level states will typically not be \ac{oi}.
Thus, we partition $\stateset_l$ according to \oicomp and require the \norm over $\left(w_h, z_h\right)$ to be the same as the \norm over each member of the partition.

\begin{figure}[tb]
  \centering
  \includestandalone[scale=0.65]{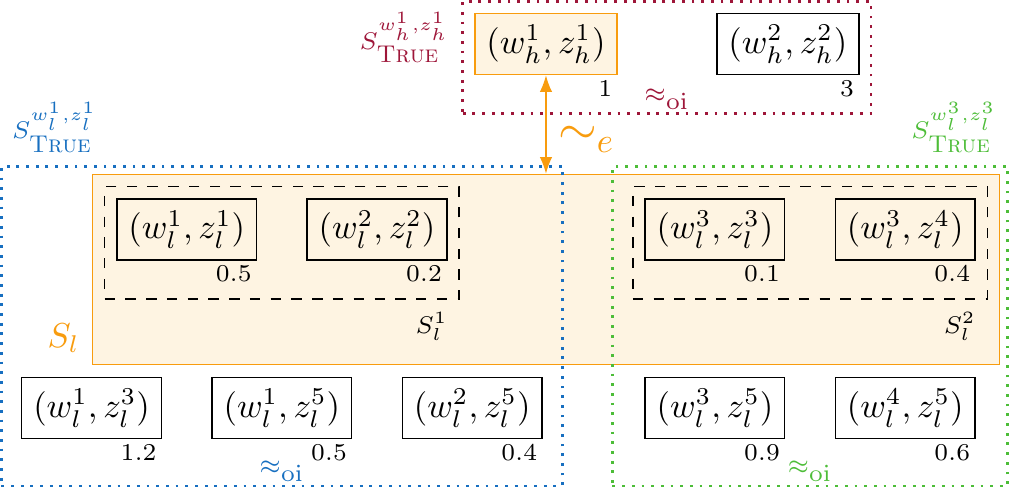}
  \caption{An example for epistemic isomorphism.}
  \label{fig:epistemic-isomorphism}
\end{figure}

\autoref{fig:epistemic-isomorphism} illustrates epistemic isomorphism.
At the top, we have the high-level state $(w_h^1, z_h^1)$ and a second high-level state $(w_h^2, z_h^2)$ that is \ac{oi} from $(w_h^1, z_h^1)$.
At the bottom, we can see that the low-level states are partitioned by $\oicomp$ into two sets, $\stateset^{w_l^1, z_l^1}_\true$ and $\stateset^{w_l^3, z_l^3}_\true$.
Horizontally aligned in the center is the set $S_l$, which is also partitioned into $S_l^1$ and $S_l^2$.
For both $S_l^1$ and $S_l^2$, the normalized weight is equal to the normalized weight of $(w_h^1, z_h^1)$, hence $(d_h, w_h^1, z_h^1)$ is epistemically isomorphic to $(d_l, S_l)$.

Having established objective and epistemic isomorphisms, we can now define a suitable notion of bisimulation:
\begin{definition}[Bisimulation]
  \label{def:bisim}
  ~ \\
  A relation
  $B \subseteq \states \times \states$
  is an \emph{$m$-bisimulation between $\left(e_h, w_h\right)$ and $\left(e_l, w_l\right)$} if $\left(\left(w_h, z_h\right), \left(w_l, z_l\right)\right) \in B$ implies that
  \begin{enumerate}
    \item \label{def:bisim:oiso}
       $\left(w_h, z_h\right) \oiso \left(w_l, z_l\right)$,
    \item \label{def:bisim:eiso}
      $\left(d_h, w_h, z_h\right) \eiso \left(d_l, \left\{ \left(w_l', z_l'\right) \mid \left(\left(w_h, z_h\right), \left(w_l', z_l'\right)\right) \in B \right\} \right)$,
    \item \label{def:bisim:executability}
      $w_h \models \exec(z_h)$ and $w_l \models \exec(z_l)$,
    \item \label{def:bisim:high-action}
      for every high-level action $a$, if $w_h, z_h \models \poss(a)$, then there is $z'_l \in \|m(a)\|^{z_l}_{e_l,w_l}$ s.t.\ $((w_h, z_h \cdot a), (w_l, z_l \cdot z_l')) \in B$,
    \item \label{def:bisim:low-action}
        for every high-level action $a$, if there is $z'_l \in \|m(a)\|^{z_l}_{e_l,w_l}$, then $w_h, z_h \models \poss(a)$ and $((w_h, z_h \cdot a), (w_l, z_l \cdot z_l')) \in B$,
      \item \label{def:bisim:high-epistemic-state}
      for every $\left(w_h', z_h'\right) \oicomp \left(w_h, z_h\right)$
      with $d_h(w_h') > 0$ and $e_h, w_h' \models \exec(z_h')$,
      there is $\left(w_l', z_l'\right) \oicomp \left(w_l, z_l\right)$
      such that 
      $\left(\left(w_h', z_h'\right), \left(w_l', z_l'\right)\right) \in B$,
    \item \label{def:bisim:low-epistemic-state}
      for every $\left(w_l', z_l'\right) \oicomp \left(w_l, z_l\right)$
      with $d_l(w_l') > 0$ and \linebreak $e_l, w_l' \models \exec(z_l')$,
      there is $\left(w_h', z_h'\right) \oicomp \left(w_h, z_h\right)$
      such that 
      $\left(\left(w_h', z_h'\right), \left(w_l', z_l'\right)\right) \in B$.
  \end{enumerate}

  We call a bisimulation $B$ \emph{definite} if $\left(\left(w_h, z_h\right), \left(w_l, z_l\right)\right) \in B$ and $\left(\left(w_h', z_h'\right), \left(w_l, z_l\right)\right) \in B$ implies $\left(w_h, z_h\right) = \left(w_h', z_h'\right)$.

We say that $\left(e_h,w_h\right)$ is bisimilar to $\left(e_l,w_l\right)$ relative to refinement mapping $m$, written $\left(e_h,w_h\right) \bisim \left(e_l,w_l\right)$, if and only if there exists a definite $m$-bisimulation relation $B$ between $\left(e_h, w_h\right)$ and $\left(e_l, w_l\right)$ such that $\left(\left(w_h, \la\ra\right), \left(w_l, \la\ra\right)\right) \in B$.

\end{definition}

The general idea of bisimulation is that two states are bisimilar if they have the same local properties (i.e., they are isomorphic) and each reachable state from the first state has a corresponding reachable state from the second state (and vice versa) such that the two successors are again bisimilar.
Here,
properties \ref{def:bisim:oiso}, \ref{def:bisim:eiso}, and \ref{def:bisim:executability} refer to static properties of $\left(w_h, z_h\right)$ and $\left(w_l, z_l\right)$.
While property \ref{def:bisim:oiso} directly establishes objective isomorphism of $\left(w_h, z_h\right)$ and $\left(w_l, z_l\right)$, property \ref{def:bisim:eiso} establishes epistemic isomorphism between $\left(w_h, z_h\right)$ and all states $\left(w_l', z_l'\right)$ that occur in $B$.
As usual in bisimulations, we also require that if we follow a high-level transition of the system, there is a corresponding low-level transition (and vice versa).
Here, such a transition may either be an action that is executed (properties \ref{def:bisim:high-action} and \ref{def:bisim:low-action}), or it may be an epistemic transition from the current state to another \ac{oi} state (properties \ref{def:bisim:high-epistemic-state} and \ref{def:bisim:low-epistemic-state}).

Our notion of bisimulation is similar to bisimulation for abstracting non-stochastic and objective basic action theories, as described by \textcite{banihashemiAbstractionSituationCalculus2017}.
In comparison, the notion of objective isomorphism (property \ref{def:bisim:oiso}) and reachable states via actions (properties \ref{def:bisim:high-action} and \ref{def:bisim:low-action}) are analogous, while epistemic isomorphism (property \ref{def:bisim:eiso}) and reachable states via observational indistinguishability (properties \ref{def:bisim:high-epistemic-state} and \ref{def:bisim:low-epistemic-state}) have no corresponding counterparts.

Given a corresponding $m$-bisimulation, we want to show that $\left(e_h, w_h\right)$ is a model of a formula $\alpha$ iff $\left(e_l, w_l\right)$ is a model of the mapped formula $m(\alpha)$.
To do so, we first show that this is true for static formulas, not considering programs.
In the second step, we will show that the high-level and low-level models induce the same program traces, which will then allow us to extend the statement to bounded formulas, which may refer to programs.
We start with static formulas:\footnote{\ifthenelse{\boolean{techreport}}{All proofs can be found in the appendix.}{All proofs can be found in the technical report.}}
\begin{theoremE}\label{thm:static-equivalence}
  Let $\left(e_h,w_h\right) \bisim \left(e_l,w_l\right)$ with definite $m$\hyp{}bisimulation $B$.
  For every static formula $\alpha$ and traces $z_h, z_l$ with $\left(\left(w_h, z_h\right), \left(w_l, z_l\right)\right) \in B$:
  \[
  e_h, w_h, z_h \models \alpha \text{ iff } e_l, w_l, z_l \models m\mleft(\alpha\mright)
  \]
\end{theoremE}
\begin{proofE}
  By structural induction on $\alpha$.
  \begin{itemize}
    \item Let $\alpha$ be an atomic formula. Then, since $\left(z_h, z_l\right) \in B$, it follows from \subdefref{def:bisim}{def:bisim:oiso}, that $\left(w_h, z_h\right) \oiso \left(w_l, z_l\right)$, and thus $w_h, z_h \models \alpha$ iff $w_l, z_l \models m \mleft(\alpha\mright)$.
    \item Let $\alpha = \beta \wedge \gamma$. The claim follows directly by induction and the semantics of conjunction.
    \item Let $\alpha = \neg \beta$. The claim follows directly by induction and the semantics of negation.
    \item Let $\alpha = \forall x.\, \beta$. The claim follows directly by induction and the semantics of all-quantification.
    \item Let $\alpha = \bel{\beta}{r}$.
      By definition, $e_h, w_h \models \bel{\beta}{r_h}$ iff
      \[
        \norm\mleft(d_h, \wh{\beta}, \wh{\true}, r_h\mright)
      \]
      Similarly, $e_l, w_l \models \bel{m(\beta)}{r_l}$ iff
      \[
        \norm\mleft(d_l, \wl{\beta}, \wl{\true}, r_l\mright)
      \]
      \underline{$r_h \leq r_l$}:
      For each $\left(w_h^i, z_h^i\right) \in \wh{\beta}$ with $d_h(w_h^i) > 0$ and $e_h, w_h^i \models \exec(z_h^i)$,
      by \subdefref{def:bisim}{def:bisim:high-epistemic-state}, there is a $\left(w_l^i, z_l^i\right)$ with $\left(\left(w_h^i, z_h^i\right), \left(w_l^i, z_l^i\right)\right) \in B$ and $\left(w_l^i, z_l^i\right) \oicomp \left(w_l, z_l\right)$.
      By \subdefref{def:bisim}{def:bisim:eiso},
      \begin{multline*}
        \left(d_h, w_h^i, z_h^i\right) \eiso \\ (d_l, \underbrace{\left\{ \left(w_l', z_l'\right) \mid \left(\left(w_h^i, z_h^i\right), \left(w_l', z_l'\right)\right) \in B \right\}}_{=: \stateset_B} )
      \end{multline*}
    Let $P$ be the partition $P = \stateset_B / \oicomp$ of $\stateset_B$. As $\left(w_l^i, z_l^i\right) \in \stateset_B$, there is a $\stateset_l^i \in P$ with $\left(w_l^i, z_l^i\right) \in \stateset_l^i$.
    By \autoref{def:eiso}:
    \begin{multline*}
      \norm(d_h, \{\left(w_h^i, z_h^i\right)\}, \wh[^i]{\true})
      \\
      =
      \norm(d_l, \stateset_l^i, \wl[^i]{\true})
    \end{multline*}
    With $\left(w_l, z_l\right) \oicomp \left(w_l^i, z_l^i\right)$, it follows that $\wl[^i]{\true} = \wl{\true}$.
    Hence:
    \begin{multline}\label{eqn:rh-leq-rl-wh-equal-sli}
      \norm(d_h, \{\left(w_h^i, z_h^i\right)\}, \wh[^i]{\true})
      \\
      =
      \norm(d_l, \stateset_l^i, \wl{\true})
    \end{multline}
    So far, we have only considered $\left(w_h^i, z_h^i\right) \in \wh{\beta}$ with $d_h(w_h^i) > 0$ and $e_h, w_h^i \models \exec(z_h^i)$.
    By definition of $\norm$, any $(w_h', z_h')$ with $d_h(w_h') = 0$ cannot add to $\norm$.
    Also, again by definition, for every $(w_h', z_h') \in \wh{\beta}$, $e_h, w_h' \models \exec(z_h')$.
    Therefore:
    \begin{multline}\label{eqn:zero-weight-states-do-not-add-anything}
      \norm(d_h, \bigcup_i \{ (w_h^i, z_h^i) \}, \wh{\true}) \\ = \norm(d_h, \wh{\beta}, \wh{\true})
    \end{multline}
    Now, as $B$ is definite, it follows for each $i \neq j$ that $\stateset_l^i \neq \stateset_l^j$ and as $P$ is a partition, $\stateset_l^i \cap \stateset_l^j = \emptyset$.
    With this and with \autoref{eqn:rh-leq-rl-wh-equal-sli} and \autoref{eqn:zero-weight-states-do-not-add-anything}, it follows that
    \begin{multline*}
      \norm(d_h, \wh{\beta}, \wh{\true})
      \\
      = \norm(d_l, \bigcup_i \stateset_l^i, \wl{\true})
    \end{multline*}

    We continue by showing the connection between all $\stateset_l^i$ and $\wl{\beta}$:
    For each $\left(w_l', z_l'\right) \in \stateset_l^i$, by definition of $\stateset_l^i$, we have $\left(w_l', z_l'\right) \oicomp \left(w_l, z_l\right)$.
    As $\left(\left(w_h^i, z_h^i\right), \left(w_l', z_l'\right)\right) \in B$, by \subdefref{def:bisim}{def:bisim:executability}, $e_l, w_l' \models \exec(z_l')$.
    Also, it follows by induction that $e_l, w_l', z_l' \models m(\beta)$.
    Thus, $\left(w_l', z_l'\right) \in \wl{\beta}$ and therefore, $\stateset_l^i \subseteq \wl{\beta}$.
    Therefore:
    \begin{multline*}
      \norm(d_l, \bigcup_i \stateset_l^i, \wl{\true})
      \\
      \leq
      \norm(d_l, \wl{\beta}, \wl{\true})
    \end{multline*}
    We summarize:
    \begin{align*}
      r_h
      &= \norm(d_h, \wh{\beta}, \wh{\true})
      \\
      &= \norm(d_l, \bigcup_i \stateset_l^i, \wl{\true})
      \\
      &\leq \norm(d_l, \wl{\beta}, \wl{\true})
      \\
      &= r_l
    \end{align*}
    Thus, $r_h \leq r_l$.


      \medskip
      \underline{$r_l \leq r_h$}:
      For each $\left(w_l^i, z_l^i\right) \in \wl{\beta}$ with $d_l(w_l^i) > 0$ and $e_l, w_l^i \models \exec(z_l^i)$,
      as $(w_l^i, z_l^i) \oicomp (w_l, z_l)$, by \subdefref{def:bisim}{def:bisim:low-epistemic-state}, there is a $\left(w_h^i, z_h^i\right)$ with
      $\left(\left(w_h^i, z_h^i\right), \left(w_l^i, z_l^i\right)\right) \in B$ and therefore, by \subdefref{def:bisim}{def:bisim:eiso},
      \begin{multline*}
        \left(d_h, w_h^i, z_h^i\right) \eiso \\ (d_l, \underbrace{\left\{ \left(w_l', z_l'\right) \mid \left(\left(w_h^i, z_h^i\right), \left(w_l', z_l'\right)\right) \in B \right\}}_{=: \stateset_B^i} )
      \end{multline*}
    Let $P$ be the partition $P = \stateset_B^i / \oicomp$ of $\stateset_B^i$. As $\left(w_l^i, z_l^i\right) \in \stateset_B^i$, there is a $\stateset_l^i \in P$ with $\left(w_l^i, z_l^i\right) \in \stateset_l^i$.
      By \autoref{def:eiso},
      \begin{multline}\label{eqn:i-norms-are-equal}
        \norm(d_l, \stateset_l^i, \wl[^i]{\true})
        \\
        =
        \norm(d_h, \{ \left(w_h^i, z_h^i\right) \}, \wh[^i]{\true})
      \end{multline}
    Now, as $\left(w_h, z_h\right) \oicomp \left(w_h^i, z_h^i\right)$, it follows that $\wh[^i]{\true} = \wh{\true}$, similarly $\wl[^i]{\true} = \wl{\true}$.
    Therefore, we can also write \autoref{eqn:i-norms-are-equal} as
    \begin{multline}\label{eqn:i-norms-over-wl-are-equal}
      \norm(d_l, \stateset_l^i, \wl{\true})
      \\
      =
      \norm(d_h, \left\{\left(w_h^i, z_h^i\right)\right\}, \wh{\true})
    \end{multline}
    Now, suppose there is $j, k$ with $j \neq k$ such that $(w_h^j, z_h^j) = (w_h^k, z_h^k)$.
    Clearly, $\stateset_B^j = \stateset_B^k$.
    Also, $(w_l^j, z_l^j) \in \wl{\beta}$ and $(w_l^k, z_l^k) \in \wl{\beta}$,
    $(w_l^j, z_l^j) \oicomp (w_l, z_l)$, $(w_l^k, z_l^k) \oicomp (w_l, z_l)$,
    and therefore also $(w_l^j, z_l^j) \oicomp (w_l^k, z_l^k)$.
    Thus, $\stateset_l^j = \stateset_l^k$.
    As \autoref{eqn:i-norms-over-wl-are-equal} holds for each $i$, it follows that
    \begin{multline}\label{eqn:sum-norms-are-equal}
      \norm(d_l, \bigcup_i \stateset_l^i, \wl{\true})
      \\
      =
      \norm(d_h, \bigcup_i \left\{\left(w_h^i, z_h^i\right)\right\}, \wh{\true})
    \end{multline}
    \\
    Let $Q = \{ \stateset_{m(\beta)}^{1}, \stateset_{m(\beta)}^{2}, \ldots \}$ be the partition of $\wl{\beta} \cap \{ \left(w_l', z_l'\right) \mid d_l(w_l') > 0 \}$ such that
    $\stateset_{m(\beta)}^{i} \subseteq \stateset_l^i$.
    With \autoref{eqn:i-norms-over-wl-are-equal}, it directly follows that
    \begin{align}\label{eqn:sum-norms-l-is-leq}
      &\norm(d_l, \bigcup_i \stateset_{m(\beta)}^{i}, \wl{\true}) \nonumber
      \\
      &\leq
      \norm(d_l, \bigcup_i \stateset_l^i, \wl{\true}) \nonumber
      \\
      &=
      \norm(d_h, \bigcup_i \left\{\left(w_h^i, z_h^i\right)\right\}, \wh{\true})
    \end{align}
    By definition of \norm, any $(w_l', z_l')$ with $d_l(w_l') = 0$ cannot add to \norm, i.e.,
    \begin{multline*}
      \norm(d_l, \bigcup_i \stateset_{m(\beta)}^i, \wl{\true}) \\ = \norm(d_l, \wl{\beta}, \wl{\true})
    \end{multline*}
    With that, \autoref{eqn:sum-norms-l-is-leq} can be written as:
    \begin{multline}\label{eqn:l-norm-is-leq-sum-h-norms}
      \norm(d_l, \wl{\beta}, \wl{\true})
      \\
      \leq
      \norm(d_h, \bigcup_i \left\{\left(w_h^i, z_h^i\right)\right\}, \wh{\true})
    \end{multline}
    Finally, as $\left(\left(w_h^i, z_h^i\right), \left(w_l^i, z_l^i\right)\right) \in B$
    and $e_l, w_l^i, z_l^i \models  m(\beta)$,
    it follows by induction that $e_h, w_h^i, z_h^i \models \beta$.
    Therefore, with $\left(w_h^i, z_h^i\right) \oicomp \left(w_h, z_h\right)$,
    we have $\left(w_h^i, z_h^i\right) \in \wh{\beta}$.
    Hence:
    \begin{multline*}
      \norm(d_l, \wl{\beta}, \wl{\true})
      \\
      \leq
      \norm(d_h, \wh{\beta}, \wh{\true})
    \end{multline*}
    Therefore $r_l \leq r_h$.

    With $r_h = r_l = r$, it follows that $e_h, w_h, z_h \models \bel{\beta}{r}$ iff $e_l, w_l, z_l \models \bel{m(\beta)}{r}$.
  \end{itemize}
\end{proofE}

Using \autoref{thm:static-equivalence}, we can show that if $\left(e_h, w_h\right)$ is bisimilar to $\left(e_l, w_l\right)$, then $\left(e_h, w_h\right)$ and $\left(e_l, w_l\right)$ induce the same traces of a program $\delta$:
\begin{lemmaE}\label{lma:bisimulation-traces}
  Let $\left(e_h,w_h\right) \bisim \left(e_l,w_l\right)$ with $m$-bisimulation $B$, $\left(\left(w_h, z_h\right), \left(w_l, z_l\right)\right) \in B$, and $\delta$ be an arbitrary program.
  \begin{enumerate}
    \item If $z_l' \in \left\| m\mleft(\delta\mright) \right\|^{z_l}_{e_l, w_l}$ is a low-level trace, then there is a high-level trace $z_h' \in \left\| \delta \right\|^{z_h}_{e_h, w_h}$ such that $z_h' = \la a_1, \ldots, a_n \ra$, $z_l' =  \la m\mleft(a_1\mright), \ldots, m\mleft(a_n\mright) \ra$, and $\left(z_h \cdot z_h', z_l \cdot z_l'\right) \in B$.
    \item If $z_h' = \la a_1, \ldots, a_n \ra \in \left\| \delta \right\|^{z_h}_{e_h, w_h}$ is a high-level trace, then there is a low-level trace $z_l' \in \left\| m\mleft(\delta\mright) \right\|^{z_l}_{e_l, w_l}$ such that $z_l' = \la m\mleft(a_1\mright), \ldots, m\mleft(a_n\mright) \ra$ and $\left(z_h \cdot z_h', z_l \cdot z_l'\right) \in B$.
  \end{enumerate}
\end{lemmaE}
\begin{proofE}[no proof end][Proof Idea]
  By structural induction on $\delta$.
  Note that for any formula that occurs in $\delta$, we can use \autoref{thm:static-equivalence} to show that it is satisfied by $(e_h, w_h, z_h)$ iff it is satisfied by $(e_l, w_l, z_l)$.
  For the same reason, the same actions are possible because $e_h, w_h, z_h \models \poss(a)$ iff $e_l, w_l, z_l \models \poss(a)$.
\end{proofE}
\begin{proofE}
  ~
  \begin{enumerate}
    \item By structural induction on $\delta$.
      \begin{itemize}
        \item Let $\delta = a$ and thus $z_l' = \la m(a) \ra$. Then, by \subdefref{def:bisim}{def:bisim:executability}, $w_h, z_h \models \poss(a)$, therefore $\la a \ra \in \left\| \delta \right\|^{z_h}_{e_h, w_h}$ and also $\left( z_h \cdot a, z_l \cdot z_l'\right) \in B$.
        \item Let $\delta = \alpha?$. From $z_l' \in \left\| m\mleft(\delta\mright) \right\|^{z_l}_{e_l, w_l}$, it directly follows that $\la z_l, m(\alpha)? \ra \in \mathcal{F}^{e_l, w_l}$, $z_l' = \la\ra$ and $e_l, w_l, z_l \models m(\alpha)$.
          By \autoref{thm:static-equivalence}, it follows that $e_h, w_h, z_h \models \alpha$.
          Thus, $\la z_h, \alpha? \ra \in \mathcal{F}^{e_h, w_h}$, and therefore, for $z_h' = \la\ra$, we can follow that $z_h' \in \left\| \delta \right\|^{z_h}_{e_h, w_h}$.
          Finally, as $z_h = z_l = \la\ra$ and $\left(z_h, z_l\right) \in B$, it follows that $\left(z_h \cdot z_h', z_l \cdot z_l'\right) \in B$.
        \item Let $\delta = \delta_1; \delta_2$. By induction, for $z_l^1 \in \|m(\delta_1)\|^{z_l}_{e_l, w_l}$, there is $z_h^1 = \la a_1, \ldots, a_k \ra \in \|\delta_1\|^{z_h}_{e_h, w_h}$ with $z_l^1 = \la m(a_1), \ldots, m(a_k) \ra$ and $\left(z_h \cdot z_h^1, z_l, \cdot z_l^1\right) \in B$.
          Let $z_l^2 \in \| m(\delta_2)\|^{z_l \cdot z_l^1}_{e_l, w_l}$.
          It follows again by induction that there is $z_h^2 = \la a_{k+1}, \ldots, a_{n} \ra \in \|\delta_2\|^{z_h \cdot z_h^1}_{e_h, w_h}$  and such that $z_l^2 = \la m(a_{k+1}), \ldots m(a_n) \ra$ and $\left(z_h \cdot z_h^1 \cdot z_h^2, z_l \cdot z_l^1 \cdot z_l^2\right) \in B$.
        \item Let $\delta = \delta_1 | \delta_2$.
          Two cases:
          \begin{enumerate}
            \item $z_l' \in \| m(\delta_1) \|^{z_l}_{e_l, w_l}$.
              Then, by induction, there is $z_h' \in \left\| \delta_1 \right\|^{z_h}_{e_h, w_h}$ with $z_h' = \la a_1, \ldots, a_n \ra$ and such that $z_l' =  \la m\mleft(a_1\mright), \ldots, m\mleft(a_n\mright) \ra$ with $\left(z_h \cdot z_h', z_l \cdot z_l'\right) \in B$.
            \item $z_l' \in \| m(\delta_2) \|^{z_l}_{e_l, w_l}$.
              Then, by induction, there is $z_h' \in \left\| \delta_2 \right\|^{z_h}_{e_h, w_h}$ with $z_h' = \la a_1, \ldots, a_n \ra$ and such that $z_l' =  \la m\mleft(a_1\mright), \ldots, m\mleft(a_n\mright) \ra$ with $\left(z_h \cdot z_h', z_l \cdot z_l'\right) \in B$.
          \end{enumerate}
      \end{itemize}
    \item By structural induction on $\delta$.
      \begin{itemize}
        \item Let $\delta = a$ and thus $z_h' = \la a \ra \in \left\| \delta \right\|^{z_h}_{e_h, w_h}$.
          Therefore, $e_h, w_h, z_h \models \poss(a)$ and thus, by \autoref{def:bisim}, there is $z_l' \in \left\| m(a) \right\|^{z_l}_{e_l, w_l}$ with $\left(z_h \cdot z_h', z_l \cdot z_l'\right) \in B$.
        \item Let $\delta = \alpha?$.
          From $z_h' \in \left\| \delta \right\|^{z_h}_{e_h, w_h}$, it directly follows that $\la z_h, a? \ra \in \mathcal{F}^{e_h, w_h}$, $z_h' = \la\ra$, and $e_h, w_h, z_h \models \alpha$.
          By \autoref{thm:static-equivalence}, it follows that $e_l, w_l, z_l \models m(\alpha)$.
          Thus, $\la z_l, m(\alpha)? \ra \in \mathcal{F}^{e_l, w_l}$, and therefore $z_l' = \la\ra \in \left\| m(\delta) \right\|^{z_l}_{e_l, w_l}$.
          Finally, as $z_h = z_l = \la\ra$ and $\left(z_h, z_l\right) \in B$, it follows that $\left(z_h \cdot z_h', z_l \cdot z_l'\right) \in B$.
        \item Let $\delta = \delta_1; \delta_2$.
          By induction, for $z_h^1 = \la a_1, \ldots, a_k \ra \in \| \delta_1 \|^{z_h}_{e_h, w_h}$, there is $z_l^1 \in \| m(\delta_1) \|^{z_l}_{e_l, w_l}$ such that $z_l^1 = \la m(a_1), \ldots, m(a_k) \ra$ with $\left(z_h \cdot z_h^1, z_l \cdot z_l^1\right) \in B$.
          Again by induction, for $z_h^2 = \la a_{k+1}, \ldots a_n \ra \in \| \delta_2 \|^{z_h \cdot z_h^1}_{e_h, w_h}$, there is $z_l^2 \in \| m(\delta_2) \|^{z_l \cdot z_l^1}_{e_l, w_l}$ such that $(z_h \cdot z_h^1 \cdot z_h^2, z_l \cdot z_l^1 \cdot z_l^2) \in B$.
        \item Let $\delta = \delta_1 | \delta_2$.
          Two cases:
          \begin{enumerate}
            \item $z_h' = \la a_1, \ldots, a_n \ra \in \| \delta_1 \|^{z_h}_{e_h, w_h}$.
              Then, by induction, there is $z_l' \in \left\| m\mleft(\delta_1\mright) \right\|^{z_l}_{e_l, w_l}$ with $z_l' = \la m\mleft(a_1\mright), \ldots, m\mleft(a_n\mright) \ra$, and $\left(z_h \cdot z_h', z_l \cdot z_l'\right) \in B$.
            \item $z_h' = \la a_1, \ldots, a_n \ra \in \| \delta_2 \|^{z_h}_{e_h, w_h}$.
              Then, by induction, there is $z_l' \in \left\| m\mleft(\delta_2\mright) \right\|^{z_l}_{e_l, w_l}$ with $z_l' = \la m\mleft(a_1\mright), \ldots, m\mleft(a_n\mright) \ra$ and $\left(z_h \cdot z_h', z_l \cdot z_l'\right) \in B$.
          \end{enumerate}
      \end{itemize}
  \end{enumerate}
\end{proofE}

Note that \autoref{lma:bisimulation-traces} would not hold if $\delta$ contained interleaved concurrency. Intuitively, this is because for a high-level program such as $a_h^1 \| a_h^2$, the only valid high-level traces would be $\la a_h^1, a_h^2\ra$ and $\la a_h^2, a_h^1\ra$, i.e., one action is completely executed before the other action is started.
On the other hand, with $m(a_h^1) = a_l^1; a_l^2$ and $m(a_h^2) = a_l^3; a_l^4$, we may obtain interleaved traces such as $\la a_l^1, a_l^3, a_l^2, a_l^4\ra$, which does not have a corresponding high-level trace.
While a limited form of concurrency could be permitted by only allowing interleaved execution of high-level actions (i.e., each $m(a)$ must be completely executed before switching to a different branch of execution), we omit this for the sake of simplicity.

With \autoref{lma:bisimulation-traces}, we can extend \autoref{thm:static-equivalence} to bounded formulas:
\begin{theoremE}\label{thm:model-mapping}
  Let $\left(e_h,w_h\right) \bisim \left(e_l,w_l\right)$ with $m$-bisimulation $B$.
  For every bounded formula $\alpha$ and traces $z_h, z_l$ with $\left(z_h, z_l\right) \in B$:
  \[
  e_h, w_h, z_h \models \alpha \text{ iff } e_l, w_l, z_l \models m\mleft(\alpha\mright)
  \]
\end{theoremE}
\begin{proofE}[no proof end][Proof Idea]
  By structural induction on $\alpha$, similarly to \autoref{thm:static-equivalence}.
  For formulas of the form $\alpha = [\delta] \beta$, it can be shown with \autoref{lma:bisimulation-traces} that they induce the same traces, which allows us to apply \autoref{thm:static-equivalence} again.
\end{proofE}
\begin{proofE}
  By structural induction on $\alpha$.
  \begin{itemize}
    \item Let $\alpha$ be an atomic formula. Then, since $\left(z_h, z_l\right) \in B$, we know that $\left(w_h, z_h\right) \oiso \left(w_l, z_l\right)$, and thus $w_h, z_h \models \alpha$ iff $w_l, z_l \models m \mleft(\alpha\mright)$.
    \item Let $\alpha = \bel{\beta}{r}$.
      Same proof as in \autoref{thm:static-equivalence}.
    \item Let $\alpha = \beta \wedge \gamma$. The claim follows directly by induction and the semantics of conjunction.
    \item Let $\alpha = \neg \beta$. The claim follows directly by induction and the semantics of negation.
    \item Let $\alpha = \forall x.\, \beta$. The claim follows directly by induction and the semantics of all-quantification.
    \item Let $\alpha = [\delta] \beta$.
      \\
      \textbf{$\Leftarrow$:}
      Let $e_h, w_h, z_h \not\models [\delta]\beta$.
      There is a finite trace $z'_h \in \|\delta\|^{z_h}_{e_h,w_h}$ with $e_h, w_h, z_h \cdot z'_h \not\models \beta$.
      By \autoref{lma:bisimulation-traces}, there is $z_l' \in \|m(\delta)\|^{z_l}_{e_l, w_l}$ with $\left(z_h \cdot z_h', z_l \cdot z_l'\right) \in B$.
      By induction, $e_l, w_l, z_l \cdot z_l' \not\models \beta$, and thus $e_l, w_l, z_l \not\models [m(\delta)]m(\beta)$.
      \\
      \textbf{$\Rightarrow$:}
      Let $\left(e_l,w_l\right) \not\models [m\mleft(\delta\mright)] m\mleft(\beta\mright)$, i.e., there is a finite trace $z_l' \in \|m(\delta)\|^{z_l}_{e_l, w_l}$ with $e_l, w_l, z_l \cdot z_l' \not\models m(\beta)$.
      By \autoref{lma:bisimulation-traces}, there is a $z_h' \in \| \delta \|^{z_h}_{e_h, w_h}$ with $\left(z_h \cdot z_h', z_l \cdot z_l'\right) \in B$.
      By induction, $e_h, w_h, z_h \cdot z_h' \not\models \beta$ and thus $e_h, w_h, z_h \not\models [\delta]\beta$.
  \end{itemize}
\end{proofE}

It directly follows that the high- and low-level models entail the same formulas after executing some program $\delta$:
\begin{corollaryE}\label{thm:mapping-satisfiability}
  Let $\left(e_h,w_h\right) \bisim \left(e_l,w_l\right)$.
  Then for any high-level Golog program $\delta$ and static high-level formula $\beta$:
  \[
    e_l,w_l \models [m\mleft(\delta\mright)] m\mleft(\beta\mright)
    \Leftrightarrow
    e_h,w_h \models [\delta]\beta
  \]
\end{corollaryE}
\begin{proofE}
  This is a special case of \autoref{thm:model-mapping} with $z_h = \la\ra, z_l = \la\ra, \alpha = [\delta]\beta$.
\end{proofE}

\subsection{Sound and Complete Abstraction}
In the previous section, we described properties of abstraction with respect to particular models $\left(e_h, w_h\right)$ and $\left(e_l, w_l\right)$.
However, we are usually more interested in the relationship between a high-level \ac{BAT} $\Sigma_h$ and a low-level \ac{BAT} $\Sigma_l$:%
\footnote{
Notice that we require the real world to have the same physical laws as
that believed by the agent, which is fairly standard.
We do not require the
agent knows everything about the real world, nor that
the agent beliefs are also true in the real world.
}
\begin{definition}[Sound Abstraction]
  We say that $\Sigma_h$ is a \emph{sound abstraction of $\Sigma_l$ relative to refinement mapping $m$} if and only if for each model $\left(e_l, w_l\right) \models \know{\Sigma_l} \wedge \Sigma_l$, there exists a model $\left(e_h, w_h\right) \models \know{\Sigma_h} \wedge \Sigma_h$ such that $\left(e_h, w_h\right) \bisim \left(e_l, w_l\right)$.
\end{definition}

We can show that conclusions by $\Sigma_h$ are consistent with $\Sigma_l$:
\begin{theoremE}\label{thm:sound-abstraction}
  Let $\Sigma_h$ be a sound abstraction of $\Sigma_l$ relative to mapping $m$.
Then, for every bounded formula $\alpha$, if $\know{\Sigma_h} \wedge \Sigma_h \models \alpha$, then $\know{\Sigma_l} \wedge \Sigma_l \models m(\alpha)$.
\end{theoremE}
\begin{proofE}
  Let $\know{\Sigma_h} \wedge \Sigma_h \models \alpha$.
  Suppose $\know{\Sigma_l} \wedge \Sigma_l \not\models m(\alpha)$,
  i.e., there is a model $\left(e_l, w_l\right)$ of $\know{\Sigma_l} \wedge \Sigma_l$ with $e_l, w_l \not\models m(\alpha)$.
  As $\Sigma_h$ is a sound abstraction of $\Sigma_l$, there is a model $\left(e_h, w_h\right)$ of $\know{\Sigma_h} \wedge \Sigma_h$ with $\left(e_h, w_h\right) \bisim \left(e_l, w_l\right)$.
  By \autoref{thm:model-mapping}, $e_h, w_h \not\models \alpha$.
  Contradiction to $\know{\Sigma_h} \wedge \Sigma_h \models \alpha$.
  Thus, $\know{\Sigma_l} \wedge \Sigma_l \models  m(\alpha)$.
\end{proofE}



\begin{figure}[tb]
  \centering
  \includestandalone[width=0.8\columnwidth]{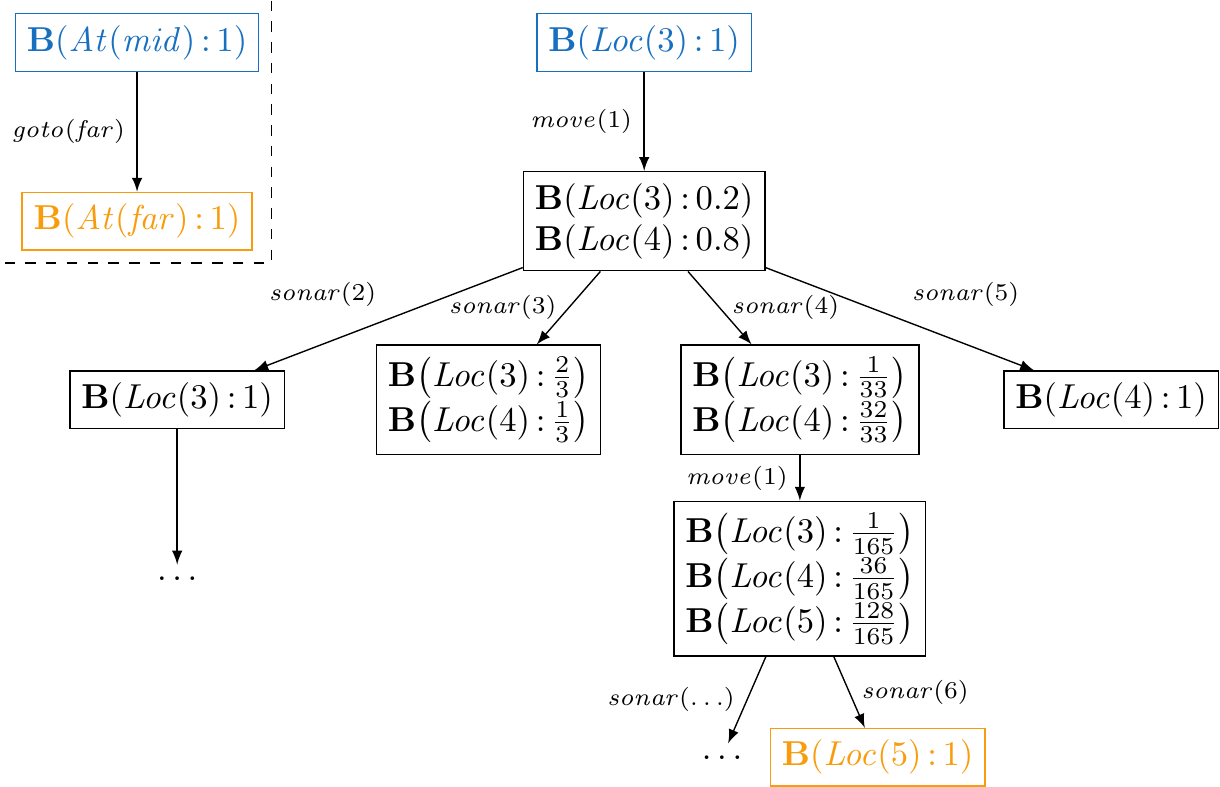}
  \caption[Bisimulation.]{
    Bisimulation for the running example, where sets of states are summarized by the belief that they entail.
}
  \label{fig:bisimulation}
\end{figure}

While a sound abstraction ensures that any entailment of the high-level \ac{BAT} $\Sigma_h$ is consistent with the low-level \ac{BAT} $\Sigma_l$, the $\Sigma_h$ may have less information than $\Sigma_l$, e.g., $\Sigma_h$ may consider it possible that some program $\delta$ is executable, while $\Sigma_l$ knows that it is not.
This leads to a second notion of abstraction:
\begin{definition}[Complete Abstraction]
  We say that $\Sigma_h$ is a \emph{complete abstraction of $\Sigma_l$ relative to refinement mapping $m$} if and only if for each model $\left(e_h, w_h\right) \models \know{\Sigma_h} \wedge \Sigma_h$,
  there exists a model $\left(e_l, w_l\right) \models \know{\Sigma_l} \wedge \Sigma_l$ such that $\left(e_h, w_h\right) \bisim \left(e_l, w_l\right)$.
\end{definition}

Indeed, if we have a complete abstraction, then $\Sigma_h$ must entail everything that $\Sigma_l$ entails:
\begin{theoremE}\label{thm:complete-abstraction}
  Let $\Sigma_h$ be a complete abstraction of $\Sigma_l$ relative to mapping $m$. Then, for every bounded formula $\alpha$, if $\know{\Sigma_l} \wedge \Sigma_l \models m(\alpha)$, then $\know{\Sigma_h} \wedge \Sigma_h \models \alpha$.
\end{theoremE}
\begin{proofE}
  Let $\know{\Sigma_l} \wedge \Sigma_l \models m(\alpha)$.
  Suppose $\know{\Sigma_h} \wedge \Sigma_h \not\models \alpha$, i.e., there is a model $\left(e_h, w_h\right)$ of $\know{\Sigma_h} \wedge \Sigma_h$ with $\left(e_h, w_h\right) \not\models \alpha$.
  As $\Sigma_h$ is a complete abstraction of $\Sigma_l$, there is a model $\left(e_l, w_l\right)$ with $e_l, w_l \models \know{\Sigma_l} \wedge \Sigma_l$ and $\left(e_h, w_h\right) \bisim \left(e_l, w_l\right)$.
  By \autoref{thm:model-mapping}, $e_l, w_l \not\models m(\alpha)$.
  Contradiction to $\Sigma_l \models m(\alpha)$.
  Thus, $\know{\Sigma_h} \wedge \Sigma_h \models \alpha$.
\end{proofE}

The strongest notion of abstraction is the combination of both: 
\begin{definition}[Sound and Complete Abstraction]
  ~ \\
  We say that $\Sigma_h$ is a \emph{sound and complete abstraction} of $\Sigma_l$ relative to refinement mapping $m$ if $\Sigma_h$ is both a sound and a complete abstraction of $\Sigma_l$ wrt $m$.
\end{definition}

\begin{theoremE}\label{thm:sound-complete-abstraction}
  Let $\Sigma_h$ be a sound and complete abstraction of $\Sigma_l$ relative to refinement mapping $m$.
  Then, for every bounded formula $\alpha$, $\know{\Sigma_h} \wedge \Sigma_h \models \alpha$ iff $\know{\Sigma_l} \wedge \Sigma_l \models m(\alpha)$.
\end{theoremE}
\begin{proofE}
  Follows directly from \autoref{thm:sound-abstraction} and \autoref{thm:complete-abstraction} .
\end{proofE}

Coming back to our example, we can show that \sigmah is indeed a sound and complete abstraction of \sigmal:
\newcommand*{\nonhfluents}{\ensuremath{\mathcal{F} \setminus \mathcal{F}_h}\xspace}
\begin{theoremE}\label{thm:sound-bat}
  \sigmah is a sound and complete abstraction of \sigmal relative to refinement mapping $m$.
\end{theoremE}
\begin{proofE}[no proof end][Proof Idea]
  Let $e_l, w_l \models \know{\sigmal} \wedge \sigmal$.
  We show by construction that there is a model $\left(e_h, w_h\right)$ with $e_h, w_h \models \know{\sigmah} \wedge \sigmah$ and $\left(e_h, w_h\right) \bisim \left(e_l, w_l\right)$.
  First, note that from $e_l, w_l \models \know{\sigmal}$, it follows that
  $d_l(w_l') > 0$ implies $w_l' \models \sigmal$.
  Let $w_h \models \sigmah$ and let $e_h$ be an epistemic state such that $d_h(w_h) = 1$ and $d_h(w_h') = 0$ for every $w_h' \neq w_h$.
  Clearly, $e_h, w_h \models \know{\sigmah} \wedge \sigmah$.
  Now, let
  \begin{multline*}
    B_0 =
    \left\{ \left(\left(w_h, \la\ra\right), \left(w_l, \la\ra\right)\right) \right\} \cup
    \\
    \left\{ \left(\left(w_h, \la\ra\right), \left(w_l', \la\ra\right)\right) \mid d_l(w_l') > 0 \right\}
  \end{multline*}
  Next, let
  \begin{multline*}
    B_{i+1} = \big\{ \left(\left(w_h', z_h' \cdot a\right), \left(w_l', z_l' \cdot z_l''\right)\right) \mid \\ \left(\left(w_h', z_h'\right), \left(w_l', z_l'\right)\right) \in B_i,
    \\
    w_h', z_h' \models \poss(a), z_l'' \in \|m(a)\|^{z_l'}_{e_l, w_l}\big\}
  \end{multline*}
  As $B$ only mentions a single high-level world $w_h$, it directly follows that $B$ is definite.
  It can be shown by induction on $i$ that $B = \bigcup_i B_i$ is an $m$-bismulation between $\left(e_h, w_h\right)$ and $\left(e_l, w_l\right)$.
  Therefore, for each $e_l, w_l \models \know{\sigmal} \wedge \sigmal$, there is a $\left(e_h, w_h\right) \models \know{\Sigma_h} \wedge \Sigma_h$ with $(e_h, w_h) \bisim (e_l, w_l)$.
  Thus, \sigmah is a sound abstraction of \sigmal.
\end{proofE}
\begin{proofE}
  ~
  \paragraph{Sound abstraction:}
  We first show that \sigmah is a sound abstraction of \sigmal:
  Let $e_l, w_l \models \know{\sigmal} \wedge \sigmal$.
  We show by construction that there is a model $\left(e_h, w_h\right)$ with $e_h, w_h \models \know{\sigmah} \wedge \sigmah$ and $\left(e_h, w_h\right) \bisim \left(e_l, w_l\right)$.

  First, note that from $e_l, w_l \models \know{\sigmal}$, it follows that
  $d_l(w_l') > 0$ implies $w_l' \models \sigmal$.
  Let $w_h \models \sigmah$ and let $e_h$ be an epistemic state such that $d_h(w_h) = 1$ and $d_h(w_h') = 0$ for every $w_h' \neq w_h$.
  Clearly, $e_h, w_h \models \know{\sigmah} \wedge \sigmah$.
  Now, let
  \begin{multline*}
    B_0 =
    \left\{ \left(\left(w_h, \la\ra\right), \left(w_l, \la\ra\right)\right) \right\} \cup
    \\
    \left\{ \left(\left(w_h, \la\ra\right), \left(w_l', \la\ra\right)\right) \mid d_l(w_l') > 0 \right\}
  \end{multline*}
  Next, let
  \begin{multline*}
    B_{i+1} = \big\{ \left(\left(w_h', z_h' \cdot a\right), \left(w_l', z_l' \cdot z_l''\right)\right) \mid \\ \left(\left(w_h', z_h'\right), \left(w_l', z_l'\right)\right) \in B_i,
    \\
                                                                                                            w_h', z_h' \models \poss(a), z_l'' \in \|m(a)\|^{z_l'}_{e_l, w_l}\big\}
  \end{multline*}
  As $B$ only mentions a single high-level world $w_h$, it directly follows that $B$ is definite.
  We show by induction on $i$ that $B = \bigcup_i B_i$ is an $m$-bismulation between $\left(e_h, w_h\right)$ and $\left(e_l, w_l\right)$.
  Let $\left(\left(w_h', z_h'\right), \left(w_l', z_l'\right)\right) \in B_i$.

  \medskip
  \noindent \textbf{Base case.}
  Note that $z_h' = z_l' = \la\ra$ by definition of $B_0$.
  We show that all criteria of \autoref{def:bisim} are satisfied:
  \begin{enumerate}
    \item By definition, $w_l' \models \sigmal$ and thus $w_l' \models \loc(x) \equivspace x = 3$.
      At the same time, $w_h' \models \sigmah$ and thus $w_h' \models \at(l) \equiv l = \midpos$.
      Therefore, for all $l$, $w_h' \models \at(l)$ iff $w_l' \models m(\at(l))$ and thus, $\left(w_h', \la\ra\right) \oiso \left(w_l', \la\ra\right)$.
    \item By definition of $e_h$, $\norm(d_h, \{ \left(w_h, \la\ra\right) \}, \wh{\true}, 1)$.
      Also, $d_l(w_l') > 0$ iff $\left(\left(w_h, \la\ra\right), \left(w_l', \la\ra\right)\right) \in B$.
      Let $\stateset_l = \{ \left(w_l', \la\ra\right) \mid d_l(w_l') > 0 \}$.
      It directly follows that for each set $\stateset_l^i$ of the partition $\stateset_l / \oicomp$,
      $\norm(d_l, \stateset_l^i, \wlfull{e_l, [\stateset_l^{(i)}]}{\true}, 1)$.
      Thus, $\left(d_h, w_h, \la\ra\right) \eiso \left(d_l, \stateset_l\right)$.
    \item As $z_h' = z_l' = \la\ra$, it directly follows that $e_h, w_h' \models \exec(z_h')$ and $e_l, w_l' \models \exec(z_l')$.
    \item Let $w_h' \models \poss(a)$.
      Then, $a = \goto(l)$ for some $l \in \{ \near, \far \}$.
      As $e_l, w_l' \models \sigmal$, it follows for each such $l$ that there is some $z_l'' \in \|m(\goto(l))\|^{\la\ra}_{e_l, w_l'}$:
      \begin{itemize}
        \item For $l = \near$, $z_l'' = \la \sonar(), \move(-1, -1), \sonar() \ra$.
        \item For $l = \far$, \newline $z_l'' = \la \sonar, \move(1, 1), \sonar(), \move(1, 1), \sonar() \ra$.
      \end{itemize}
      By definition of $B_{i+1}$, we obtain $\left(\left(w_h', z_h' \cdot a\right), \left(w_l', z_l' \cdot z_l''\right)\right) \in B$.
    \item Let $z_l'' \in \|m(a)\|^{z_l'}_{e_l, w_l}$.
      Clearly, $a = \goto(l)$ for some $l \in \{ \near, \far \}$.
      By definition of \sigmah, it directly follows that $e_h, w_h', z_h' \models \poss(a)$.
      By definition of $B_{i+1}$, it also follows that $\left(\left(w_h', z_h' \cdot a\right), \left(w_l', z_l' \cdot z_l''\right)\right) \in B$.
    \item Let $\left(w_h'', z_h''\right) \oicomp \left(w_h', z_h'\right)$ with $d_h(w_h'') > 0$ and $e_h, w_h'' \models \exec(z_h'')$.
      As $d_h(w_h'') = 0$ for every $w_h'' \neq w_h'$ and $z_h' = \la\ra$, it directly follows that $\left(w_h'', z_h''\right) = \left(w_h', z_h'\right)$
      and thus
      $\left(\left(w_h'', z_h''\right), \left(w_l', z_l'\right)\right) \in B$.
    \item Let $\left(w_l'', z_l''\right) \oicomp \left(w_l', z_l'\right)$ with $d_l(w_l'') > 0$ and $e_h, w_l'' \models \exec(z_l'')$.
      Clearly, $z_l'' = z_l' = \la\ra$.
      By definition of $B_0$, $\left(\left(w_h, \la\ra\right), \left(w_l'', \la\ra\right)\right) \in B$.
  \end{enumerate}
  \medskip
  \noindent \textbf{Induction step.}
  \begin{enumerate}
    \item Let $z_h' = z_h'' \cdot a$ and $z_l' = z_l'' \cdot z_l'''$ for some $z_l''' \in \|m(a)\|^{z_l''}_{e_l, w_l'}$.
      By construction, $\left(\left(w_h', z_h''\right), \left(w_l', z_l''\right)\right) \in B$.
      By induction, $\left(w_h', z_h''\right) \oiso \left(w_l', z_l''\right)$.
      Furthermore, $w_h' \models \sigmah$ and $w_l' \models \sigmal$.
      As before, $a = \goto(l)$ with $l \in \{ \near, \far \}$.
      As $w_h' \models \sigmah$, for every $l'$, $e_h, w_h', z_h' \models \at(l')$ iff $l' = l$.
      Similarly, by definition of $m$ and $\sigmal$, for every $l'$, $e_l, w_l', z_l' \models m(\at(l'))$ iff $l' = l$.
      Thus, $\left(w_h', z_h'\right) \oiso \left(w_l', z_l'\right)$.
    \item Suppose
      \begin{multline*}
        \left(d_h, w_h', z_h'\right) \not\eiso \\ (d_l, \underbrace{\left\{ \left(w_l'', z_l''\right) \mid \left(\left(w_h', z_h'\right), \left(w_l'', z_l''\right)\right) \in B \right\}}_{=:\stateset_B} )
      \end{multline*}
      First, note that $\norm(d_h, \{ \left(w_h', z_h'\right) \}, \wh[']{\true}, 1)$.
      Therefore, there is a $\stateset_l^i \in \stateset_B / \oicomp$ and $(w_l^i, z_l^i) \in \stateset_l^i$ where $\norm(d_l, \stateset_l^i, \wl[^i]{\true}, 1) \neq 1$,
      i.e., there is $\left(w_l'', z_l''\right) \in \wl[^i]{\true} \setminus \stateset_l^i$ with $d_l(w_l'') \times l^*(w_l'', z_l'') > 0$.
      It follows that $\left(w_l'', z_l''\right) \oicomp \left(w_l^i, z_l^i\right)$ for some $\left(w_l^i, z_l^i\right) \in \stateset_l^i$.
      But then, by definition of \sigmal, $z_l''$ is the same as $z_l^i$, except a possibly different second parameter of each $\move(x, y)$ action.
      Also, $z_l^i = z_l^{i,1} \cdot z_l^{i,2}$, where $z_l^{i,2} \in \| m(a) \|^{z_l^{i,1}}_{e_l, w_l^i}$ for some action $a$.
      As $z_l'' \oisim[w_l''] z_l^i$, it follows that $z_l'' = z_l^{''1} \cdot z_l^{''2}$ with $z_l^{''2} \in \| m(a) \|^{z_l^{''1}}_{e_l,w_l''}$
      and $z_l^{''1} \oisim[w_l''] z_l^{i,1}$.
      But then, by induction, there is some $(w_h'', z_h^{''1})$ such that $((w_h'', z_h^{''1}), (w_l'', z_l^{''1})) \in B$,
      and therefore, by definition of $B$, also $((w_h'', z_h^{''1} \cdot a), (w_l'', z_l'')) \in B$.
      Contradiction to $\left(w_l'', z_l''\right) \in \wl[^i]{\true} \setminus \stateset_l^i$.
      It follows:
      \begin{multline*}
        \left(d_h, w_h', z_h'\right) \eiso \\ (d_l, \underbrace{\left\{ \left(w_l'', z_l''\right) \mid \left(\left(w_h', z_h'\right), \left(w_l'', z_l''\right)\right) \in B \right\}}_{=:\stateset_B} )
      \end{multline*}
    \item $w_h' \models \exec(z_h')$ and $w_l' \models \exec(z_l')$ directly follows by construction of $B$.
    \item Let $w_h', z_h' \models \poss(a)$.
      Then, $a = \goto(l)$ for some $l \in \{ \near, \far \}$.
      As $e_l, w_l' \models \sigmal$, it follows for each such $l$ that there is some $z_l'' \in \|m(\goto(l))\|^{z_l'}_{e_l, w_l'}$.
      By definition of $B_{i+1}$, it also follows that $\left(\left(w_h', z_h' \cdot a\right), \left(w_l', z_l' \cdot z_l''\right)\right) \in B$.
    \item Let $z_l'' \in \|m(a)\|^{z_l'}_{e_l, w_l}$.
      Clearly, $a = \goto(l)$ for some $l \in \{ \near, \far \}$.
      By definition of \sigmah, it directly follows that $e_h, w_h', z_h' \models \poss(a)$.
      By definition of $B_{i+1}$, it also follows that $\left(\left(w_h', z_h' \cdot a\right), \left(w_l', z_l' \cdot z_l''\right)\right) \in B$.
    \item Let $\left(w_h'', z_h''\right) \oicomp \left(w_h', z_h'\right)$ with $d_h(w_h'') > 0$ and $e_h, w_h'' \models \exec(z_h'')$.
      As $d_h(w_h'') = 0$ for every $w_h'' \neq w_h'$, it follows that $w_h'' = w_h'$.
      Furthermore, by definition of \sigmah, $z_h'' \oisim[w_h''] z_h'$ iff $z_h'' = z_h'$, therefore
       $\left(w_h'', z_h''\right) = \left(w_h', z_h'\right)$ and thus
      $\left(\left(w_h'', z_h''\right), \left(w_l', z_l'\right)\right) \in B$.
    \item Let $\left(w_l'', z_l''\right) \oicomp \left(w_l', z_l'\right)$ with $d_l(w_l'') > 0$ and $e_h, w_l'' \models \exec(z_l'')$.
      As $z_l'' \oisim[w_l''] z_l'$, the trace $z_l''$ must consist of the same actions as $z_l'$, except for a possibly different second parameter in each $\move(x, y)$.
      Furthermore, as \sigmah only contains the action \goto, the trace $z_l'$ only consists of mapped \goto actions, i.e., $z_l' \in \| m(\goto(l_1)); \ldots; m(\goto(l_n))\|^{\la\ra}_{e_l, w_l'}$
      We can split $z_l' = z_l^{'1} \cdot z_l^{'2}$ such that $z_l^{'2} \in \|m(\goto(l_n))\|^{z_l^{'1}}_{e_l, w_l'}$.
      Then, because of $z'' \oisim[w_l''] z_l'$, we can also split $z_l''$ such that $z_l'' = z_l^{''1} \cdot z_l^{''2}$, $z_l^{''1} \oisim[w_l'']  z_l^{'1}$ with $z_l^{''2} \in \|m(\goto(l_n'))\|^{z_l^{''1}}_{e_l, w_l''}$.
      By induction, $((w_l', z_l^{'1}), (w_l'', z_l^{''1})) \in B$.
      Finally, as $z_l^{''2} \in \|m(\goto(l_n'))\|^{z_l^{''1}}_{e_l, w_l''}$, it follows that $\left(\left(w_h', z_h'\right), \left(w_l'', z_l''\right)\right) \in B$ by definition of $B_{i}$.
  \end{enumerate}
  We conclude that $B$ is an $m$-bisimulation between $\left(e_h, w_h\right)$ and  $\left(e_l, w_l\right)$. Therefore, $\left(e_h, w_h\right) \bisim \left(e_l, w_l\right)$, and therefore $\sigmah$ is a sound abstraction of $\sigmal$.

  \paragraph{Complete abstraction:}
  We now show that \sigmah is a complete abstraction of \sigmal:
  Let $e_h, w_h \models \know{\sigmah} \wedge \sigmah$.
  We show by construction that there is a model $\left(e_l, w_l\right)$ with $e_l, w_l \models \know{\sigmal} \wedge \sigmal$
  and $\left(e_h, w_h\right) \bisim \left(e_l, w_l\right)$.
  First, note that from $e_h, w_h \models \know{\sigmah}$ it follows that $d_h(w_h') > 0$ implies $w_h' \models \sigmal$.
  Now, for each $w_h^i$ with $d_h(w_h^i) > 0$, let $w_l^i$ be a world with $w_l^i \models \sigmal$ and such that $w_l^i$ is like $w_h^i$ for the high-level fluents, i.e., for every $F \not\in \mathcal{F}_l$ and every $z \in \traces$, $w^i[F, z] = w_h^i[F, z]$.
  Thus, $w^i$ is exactly like $w_h^i$ for every fluent not mentioned in \sigmal.
  We set $e_l(w_l^i) = e_h(w_h^i)$ and $e_l(w_l') = 0$ for every other world.
  Clearly, $e_l, w_l^i \models \know{\sigmal} \wedge \sigmal$.
  Now, let:
  \[
  B_0 = \{ \left(\left(w_h^i, \la\ra\right), \left(w_l^i, \la\ra\right)\right) \mid e_h(w_h^i) > 0 \}
  \]
  As before:
  \begin{multline*}
    B_{i+1} = \big\{ \left(\left(w_h', z_h' \cdot a\right), \left(w_l', z_l' \cdot z_l''\right)\right) \mid \\ \left(\left(w_h', z_h'\right), \left(w_l', z_l'\right)\right) \in B_i,
    \\
                                                                                                            w_h', z_h' \models \poss(a), z_l'' \in \|m(a)\|^{z_l'}_{e_l, w_l}\big\}
  \end{multline*}
  As each $w_l^i$ is like $w_h^i$, it follows that $B$ is definite.
  We can again show by induction on $i$ that $B$ is an $m$-bisimulation between $\left(e_h, w_h\right)$ and $\left(e_l, w_l\right)$. Therefore, $\left(e_h, w_h\right) \bisim \left(e_l, w_l\right)$ and thus, $\sigmah$ is a complete abstraction of $\sigmal$.
\end{proofE}

\autoref{fig:bisimulation} shows an exemplary bisimulation for the running example.
The single transition for \goto of the high-level \ac{BAT} is shown on the left.
The agent knows that it is initially in the middle and after doing $\goto(\far)$, it is far away from the wall.
Some corresponding transitions of the low-level \ac{BAT} are shown on the right:
Initially, the agent knows that it is at $\loc(3)$, which is a bisimilar state to the initial high-level state (blue).
Eventually, it reaches a state where it knows that it is at $\loc(5)$, which is again a bisimilar state to the corresponding high-level state (orange).

With \autoref{thm:sound-complete-abstraction}, it follows that both \acp{BAT} entail the same (mapped) formulas.
Therefore, we can use \sigmah for reasoning and planning, e.g., we may write a high-level \golog program in terms of \sigmah and then use a classical \golog interpreter to find a ground action sequence that realizes the program.
To continue the example, we may write a very simple abstract program $\delta_h$ that first moves to the wall if necessary and then moves back:
\begin{algorithmic}
  \IIf{$\neg \at(\near)$}
    $\goto(\near)$
  \IEndIf;
  $\goto(\far)$
\end{algorithmic}
If the robot is initially not near the wall (as in our example), the following sequence is a realization of the program:
\[
  \la \goto(\near), \goto(\far) \ra 
\]
This high-level trace is much simpler than the trace of the low-level program shown in \autoref{eqn:low-level-trace}.
At the same time, as \sigmah is a sound and complete abstraction of \sigmal, the two traces end in bisimilar states.
Hence, for execution, this sequence may be translated to \sigmal by applying the refinement mapping $m$ and the translated program then takes care of noisy sensors and actuators.

\section{Conclusion}\label{sec:conclusion}
In this paper, we have presented a framework for abstraction of probabilistic dynamic domains.
More specifically, in a first step, we have defined a transition semantics for \golog programs with noisy actions based on \ds, a variant of the situation calculus with probabilistic belief.
We have then defined a suitable notion of bisimulation in the logic that allows the abstraction of noisy robot programs in terms of a refinement mapping from an abstract to a low-level basic action theory.
As seen in the example, this abstraction method allows to obtain a significantly simpler high-level domain, which can be used for reasoning or high-level programming without the need to deal with stochastic actions.
Furthermore, for a user, the resulting programs and traces are much easier to understand, because they do not contain noisy sensors and actuators and are often much shorter.

While abstractions need to be manually constructed, future work may explore abstraction generation algorithms based on \cite{holtzenSoundAbstractionDecomposition2018,belleAbstractingProbabilisticModels2020}.
A further extension might be to provide conditions under which we can modify the low-level program, e.g., with new sensors with different error profiles, without modifying the high-level program.

Interestingly, as the logics \ds and \es are fully compatible for non-probabilistic formulas not mentioning noisy actions \cite{belleReasoningProbabilitiesUnbounded2017} and abstraction allows to get rid of probabilistic formulas and noisy actions, we may construct \es programs that are sound and complete abstractions of \ds programs.
This is a step towards cognitive robotics as envisioned by Reiter~\cite{levesqueCognitiveRobotics2008}, where the classical non-probabilistic situation calculus machinery may prove entirely sufficient to define the behavior and termination of real-world robots.

\pagebreak
\balance


\begin{acks}
Till was partly supported by the Deutsche Forschungsgemeinschaft (DFG, German Research Foundation) -- 2236/1 and the EU ICT-48 2020 project TAILOR (No.~952215).
Part of this work was created during a research visit of Till at the University of Edinburgh, which was funded by the German Academic Exchange Service (DAAD).
Vaishak was partly supported by a Royal Society University Research Fellowship, UK, and partly supported by a grant from the UKRI Strategic Priorities Fund, UK to the UKRI Research Node on Trustworthy Autonomous Systems Governance and Regulation (EP/V026607/1, 2020–2024).
\end{acks}

\bibliographystyle{ACM-Reference-Format}
\bibliography{zotero}

\clearpage

\ifthenelse{\boolean{techreport}}{
\section*{Proofs}
\printProofs
}{}

\end{document}